\documentclass[letterpaper,10pt,conference]{ieeeconf}
\usepackage[T1]{fontenc}
\let\labelindent\relax %
\usepackage{amsmath,amssymb,graphicx,booktabs,array,multirow,xcolor,enumitem}
\definecolor{mimicallorange}{HTML}{D97706}
\definecolor{mimicallgold}{HTML}{FF941A}
\definecolor{mimicallblue}{HTML}{4C6FFF}
\definecolor{oursrowbg}{HTML}{EAF3FC}
\definecolor{concatrowbg}{HTML}{FFF6DC}
\definecolor{attentionrowbg}{HTML}{EEF7E9}
\definecolor{referencepurple}{HTML}{7C3AED}
\usepackage{cite}
\usepackage{placeins}
\makeatletter
\let\NAT@parse\undefined
\makeatother
\usepackage[colorlinks=true,linkcolor=referencepurple,citecolor=referencepurple,
            urlcolor=referencepurple,filecolor=referencepurple,bookmarksopen=true]{hyperref}
\makeatletter
\newcommand{\highlighttablerow}[1]{\noalign{\begingroup\color{#1}%
  \hrule height\dimexpr\ht\@arstrutbox+\dp\@arstrutbox\relax
  \kern-\dimexpr\ht\@arstrutbox+\dp\@arstrutbox\relax\endgroup}}
\newcommand{\highlightoursrow}{\highlighttablerow{oursrowbg}}
\long\def\@makecaption#1#2{%
  \vskip5pt\begingroup\small
  \setbox\@tempboxa\hbox{#1. #2}%
  \ifdim\wd\@tempboxa>\hsize #1. #2\par
  \else\hbox to\hsize{\hfil\box\@tempboxa\hfil}\fi
  \endgroup\vskip5pt}
\makeatother
\DeclareRobustCommand{\method}{\textbf{\textcolor{mimicallgold}{Mimic}\textcolor{mimicallblue}{All}}}
\DeclareRobustCommand{\methodtitle}{\raisebox{-0.12em}{\includegraphics[height=1.25em,trim=279bp 146bp 284bp 92bp,clip]{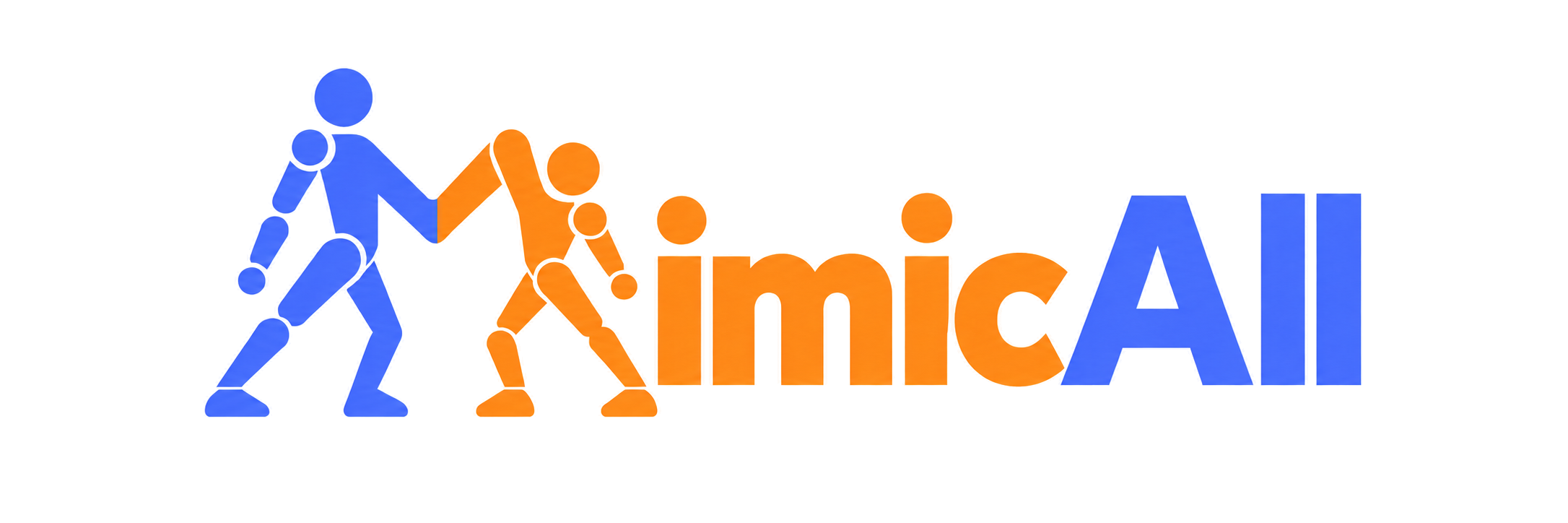}}}
\newcommand{\mask}{m^{\mathrm{act}}}

\newcommand{\TokenSubsetCorrect}{61.7}
\newcommand{\TokenSubsetZero}{33.7}
\newcommand{\TokenSubsetWrong}{47.0}
\newcommand{\TokenSubsetWrongGain}{14.6}
\newcommand{\TokenSubsetWrongLow}{13.6}
\newcommand{\TokenSubsetWrongHigh}{15.7}

\newcommand{\TokenSubsetN}{11136}
\newcommand{\TokenSubsetCorrectCount}{6867}
\newcommand{\TokenSubsetZeroCount}{3755}
\newcommand{\TokenSubsetWrongCount}{5238}

\title{\methodtitle{}: Structured Embodiment Conditioning\\for Cross-Humanoid Whole-Body Motion Tracking}
\newif\ifanonymous
\anonymousfalse %
\ifanonymous
  \author{Anonymous Authors}
  \hypersetup{pdfauthor={}}
\else
  \IEEEoverridecommandlockouts
  \author{%
\authorblockN{Jia Li$^{1}$, Lingyun Xu$^{2}$, Yitang Li$^{3}$ and Hao Zhao$^{3,4,\dagger}$\\[2pt]}
\authorblockA{%
\small
$^{1}$Huazhong University of Science and Technology\hspace{12pt}$^{2}$Peking University\\[2.25pt]
$^{3}$Tsinghua University\hspace{12pt}$^{4}$Roboparty Lab\\[4.5pt]
$^{\dagger}$Corresponding author.}
}
\hypersetup{pdfauthor={Jia Li, Lingyun Xu, Yitang Li, Hao Zhao}}

\fi
\makeatletter
\let\IEEEaftertitletext\@IEEESAVECMDIEEEaftertitletext
\IEEEaftertitletext{%
  \begin{minipage}{\textwidth}
  \centering
  \includegraphics[width=\textwidth]{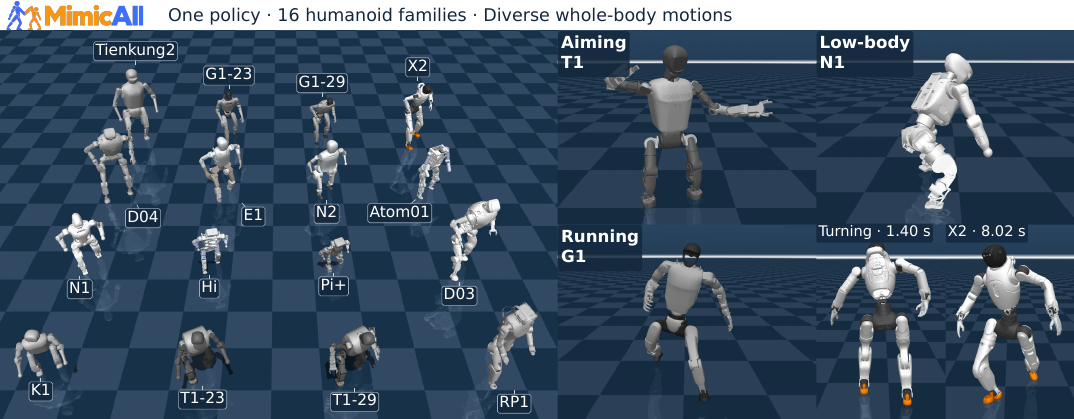}
  \def\@captype{figure}
  \caption{\textbf{One policy across diverse humanoids and motions.} 
  \method{} learns a shared whole-body tracking policy across 4,096 body variants from 16 humanoid families.
  Retargeting 64 LAFAN1 motion takes to these bodies yields approximately 262K body--motion trajectories totaling 2.93 billion reference frames.
  Demonstrations include simulated executions illustrating aiming, low-body movement, running and turning across humanoids. 
  }
  \label{fig:teaser}
  \end{minipage}\vspace{8pt}%
}
\makeatother
\begin{document}
\maketitle
\thispagestyle{empty}\pagestyle{empty}
\begin{abstract}
Humanoid motion trackers reproduce diverse behaviors, but sharing this capability across robot bodies requires a policy to account for their different structures, morphologies and dynamics.
We present \method{}, a reinforcement-learning framework that uses embodiment information to learn a shared whole-body tracking policy across heterogeneous humanoids.
These embodiment properties are represented by a structured embodiment descriptor.
An embodiment encoder produces global and joint-aligned body conditions, which the policy combines with observation history and body-matched motion references to predict joint-target offsets.
To support training across many bodies, we integrate physics-consistent embodiment randomization, body-calibrated retargeting and heterogeneous simulation through a shared joint interface.
Windowed reference streaming and body-wise encoding reuse support joint training across 4,096 variants from 16 humanoid families.
We propose a standardized benchmark spanning 464 bodies and five body-distribution strata to evaluate motion generalization, within-family embodiment shifts and unseen-family transfer.
Descriptor ablations and comparisons of four conditioning mechanisms show that embodiment information improves motion completion even with simple concatenation, while self-attention achieves the lowest global position errors.
With the same number of training transitions, shared training achieves 99.4\% thirty-second success on randomized G1 bodies in 100STYLE, compared with 42.9\% for a single-body specialist.
With Flat-MLP encoding and concatenation, continued training on a motion subset excluding obstacle categories achieves 98.3\% and 93.5\% thirty-second tracking success on 100STYLE for sampled training bodies and out-of-distribution body variants within seen families, respectively.
Across 464 bodies and an eight-motion LAFAN1 subset excluding ground and get-up sequences, paired interventions reduce Self-attention's thirty-second tracking success from \TokenSubsetCorrect{}\% to \TokenSubsetWrong{}\% when body-matched conditions are replaced with mismatched conditions, demonstrating a closed-loop benefit from embodiment information.

\end{abstract}

\section{Introduction}

Whole-body motion tracking enables humanoid controllers to reproduce diverse reference motions \cite{peng2018deepmimic,luo2023phc,luo2025sonic,chen2026holomotion}. Most existing trackers are trained for a particular robot body, yet transferring the same behavior to another body changes the joint configurations, contact patterns and actuator responses needed to execute it. A scalable cross-embodiment tracker must therefore reuse motion knowledge across embodiments while adapting control to the structure, morphology and dynamics of the executing body.

Prior cross-embodiment work has shown that control policies can share knowledge across different robots \cite{chen2018hcp,gupta2022metamorph,xue2026xhugwbc,peng2026eagle}.
Whole-body reference tracking makes the role of embodiment particularly concrete: observations describe the current execution, but joint connectivity, geometry and actuation properties determine how the body responds to a control command.
A shared policy must combine both to follow a changing full-body target.  
This raises two questions: which embodiment information should represent the executing body and help a shared policy track these references, and how should it interact with joint-wise control?  

\begin{figure*}[!t]
\centering\includegraphics[width=\textwidth]{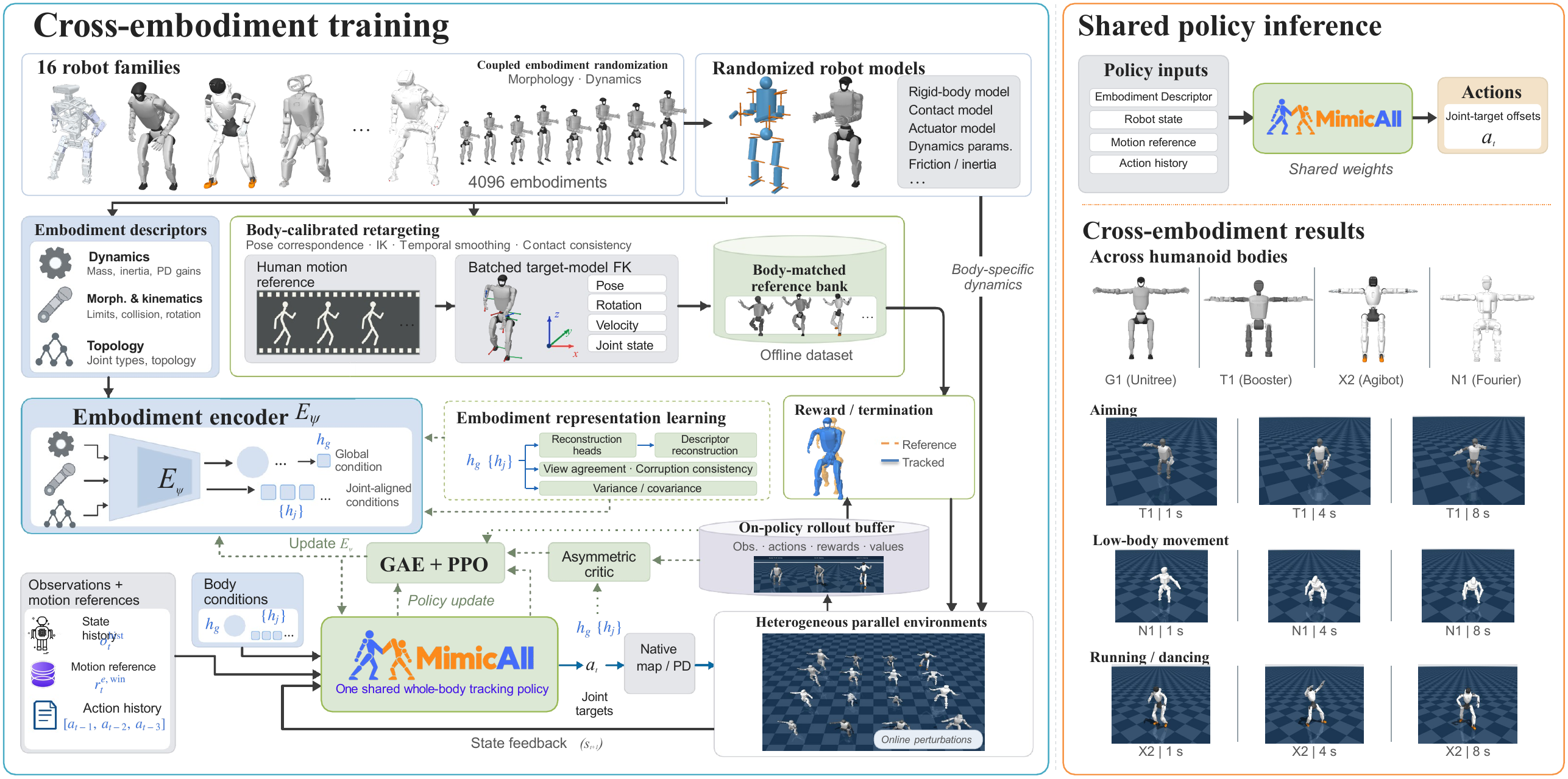}
\caption{\textbf{Cross-embodiment training and shared-policy inference.} Each randomized robot model supplies its embodiment descriptor, retargeting geometry and simulated dynamics. The encoder maps the descriptor to global and joint-aligned body conditions. A shared policy combines them with observations and body-matched references to produce joint-target offsets. Solid arrows denote forward data and control flow; dashed arrows denote training dependencies and parameter updates.}
\label{fig:pipeline}
\end{figure*}

We present \method{}, a reinforcement-learning framework that conditions a shared whole-body tracking policy on explicit embodiment information.
A structured embodiment descriptor represents each body's structure, morphology and dynamics.
An embodiment encoder converts it into global and joint-aligned body conditions: global context describes the body as a whole, while joint features retain their correspondence to action channels.
The policy combines these conditions with observations and body-matched motion references to predict joint-target offsets (Fig.~\ref{fig:pipeline}).
This formulation lets us vary descriptor content, encoder design and the conditioning mechanism without changing the tracking task.

Training across many bodies also requires a consistent path from body generation to control.
Our infrastructure couples physics-consistent embodiment randomization, body-calibrated retargeting and target-model forward kinematics, so that descriptors, reference motions and simulated dynamics refer to the same generated body.
A shared semantic joint interface supports heterogeneous simulation of robot families with different structures, morphologies and dynamics within a common training runtime.
Windowed reference streaming reduces GPU memory requirements by keeping only active reference windows resident rather than the full motion bank, making large-scale body-matched reference libraries practical for training.
Meanwhile, body-wise encoding reuse avoids redundant computation of encoding the same descriptor.
With this infrastructure, we jointly train a single tracking policy across 4,096 body variants from 16 humanoid families.

Our experiments distinguish the benefits of embodiment information from those of the conditioning mechanism.
Within seen families, embodiment descriptor conditioning improves motion completion over Mask-only even with Flat-MLP encoding and simple concatenation, whereas Self-attention achieves the lowest global position errors among the evaluated configurations.
Paired condition interventions show that body-matched information improves closed-loop tracking (Sec.~\ref{sec:rq2}).
Shared training substantially improves robustness to body variation over single-body specialization (Sec.~\ref{sec:rq1}), yet a clear gap remains between within-family generalization and unseen-family transfer (Sec.~\ref{sec:rq3}).

\noindent We make three contributions:
\begin{enumerate}[leftmargin=*, labelindent=0pt, itemsep=1pt, parsep=0pt]
\item We introduce \method{}, a scalable cross-embodiment training infrastructure for humanoid motion tracking that keeps randomized robot models, embodiment descriptors and body-matched motion references consistent, with reference streaming and encoding reuse supporting joint training across thousands of body variants.
\item We train a shared embodiment-conditioned whole-body tracker across heterogeneous humanoids and conduct a controlled study of descriptor content, encoder design and four conditioning mechanisms, identifying distinct benefits for motion completion, robustness to body variations and global tracking accuracy.
\item We establish a 464-embodiment benchmark with five embodiment-distribution strata to distinguish within-family body variation from unseen-family transfer, complemented by paired condition interventions, single-body specialist comparisons and target-robot fine-tuning.
\end{enumerate}
\begin{figure*}[t]
\centering\includegraphics[width=\textwidth]{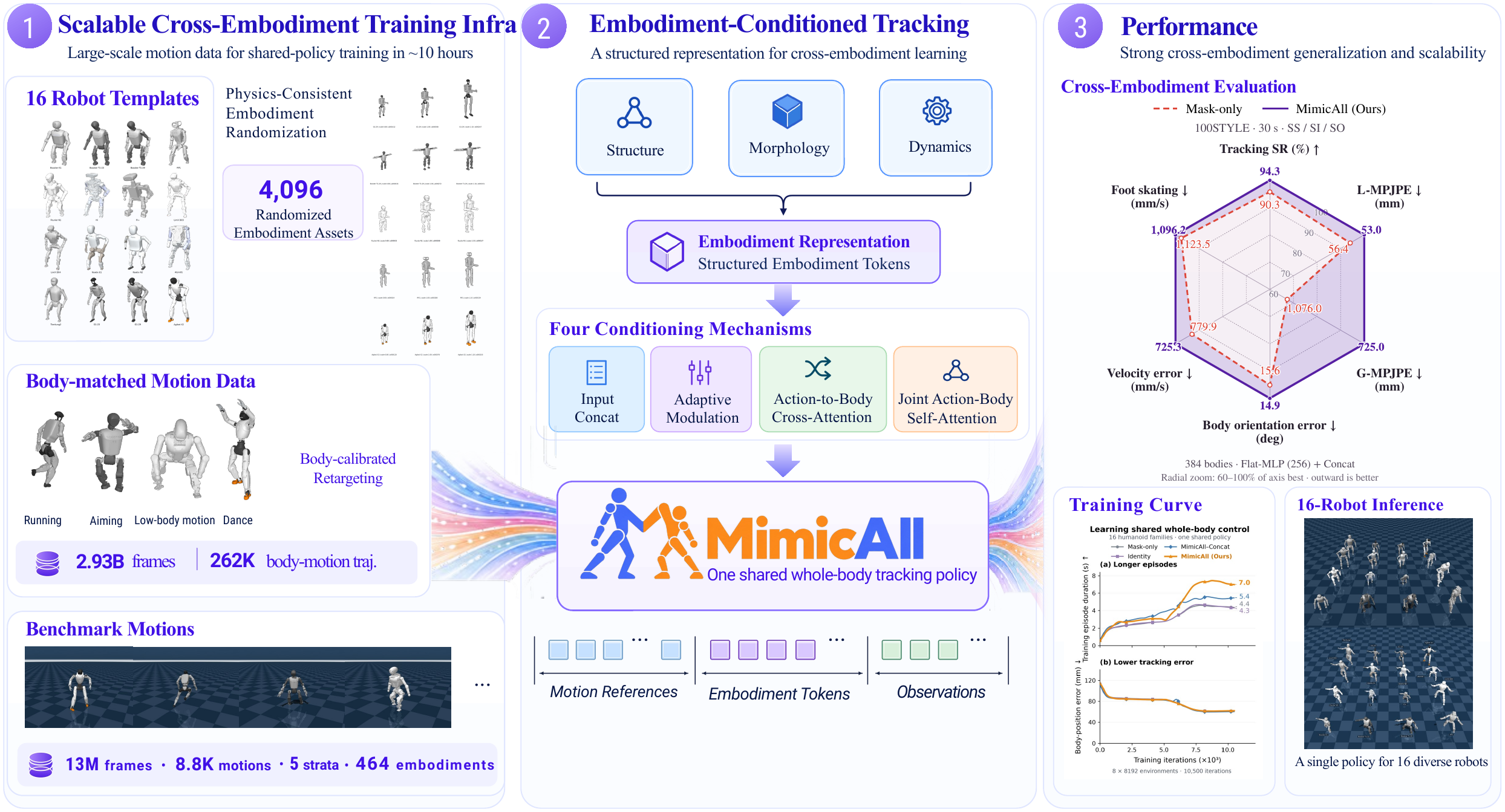}
\caption{\textbf{Overview of \method{}}. Left: coupled randomization expands 16 robot templates into 4,096 body variants; body-calibrated retargeting constructs their motion-reference bank. Center: structure, morphology and dynamics are encoded into global and joint-aligned body conditions, integrated through four conditioning variants. Right: a shared policy tracks diverse motions across heterogeneous humanoids. On held-out 100STYLE motions, Flat-MLP encoding with concatenation improves motion completion and reduces pose, velocity and foot-skating errors relative to Mask-only across SS/SI/SO bodies. Training curves and multi-robot rollouts illustrate the learning process and the resulting cross-embodiment tracking capability.
}
\label{fig:overview}
\end{figure*}

\section{Related Work}
\paragraph{Humanoid motion tracking}
Reference-conditioned controllers learn to reproduce whole-body reference motions through closed-loop control. DeepMimic and PHC establish effective learning-based tracking of simulated characters, while SONIC and HoloMotion extend the scale and diversity of humanoid tracking \cite{peng2018deepmimic,luo2023phc,luo2025sonic,chen2026holomotion}. We build on this control formulation and the MimicLite training stack \cite{robotparty2026mimiclite}, but study a different source of variation: the body executing the reference changes alongside the motion. A shared tracker must therefore use motion information without assuming one fixed kinematic structure or set of actuator responses.

\paragraph{Control across embodiments}
XHugWBC combines physics-consistent morphology randomization, semantic state--action alignment and topology-aware policy learning, without providing explicit numerical morphology and dynamics descriptors to its actor \cite{xue2026xhugwbc}. EAGLE alternates specialist learning with generalist distillation \cite{peng2026eagle}. These works demonstrate shared humanoid control under locomotion and posture commands. For full-body reference tracking, VENOM learns from multi-humanoid expert trajectories, Any2Any adapts a pretrained tracker through kinematic alignment and parameter-efficient fine-tuning, and X-WBC combines a shared temporal backbone with robot-specific input and output modules \cite{padmanabhan2026venom,yang2026any2any,zhang2026xwbc}. Our focus is explicit embodiment conditioning in a jointly trained tracker: generated bodies share policy parameters and a semantic action head, while their physical properties enter through global and joint-aligned conditions. This formulation also lets us distinguish robustness within known families from transfer to new families.

\paragraph{Embodiment representations and conditioning}
Hardware Conditioned Policies encode robot properties as policy inputs; MetaMorph combines morphology and proprioception in modular tokens \cite{chen2018hcp,gupta2022metamorph}. ModuMorph uses hypernetworks to generate morphology-dependent control parameters, Body Transformer incorporates body structure through masked attention, and URMA shares a locomotion controller through morphology-agnostic encoders and decoders \cite{xiong2023modumorph,sferrazza2024bot,bohlinger2025urma}. These approaches motivate explicit representations of the executing body. We investigate their role in whole-body tracking by separating three choices: descriptor fields, encoder architecture and interaction with the control policy.%
\paragraph{Retargeting and reference consistency}
GMR studies the importance of retargeted motion quality for downstream tracking, while OmniRetarget preserves interaction relationships during motion transfer \cite{araujo2025gmr,yang2025omniretarget}. For tracking with robot-space references, increasing body diversity requires motion references that remain consistent with each target robot's kinematics. Our infrastructure connects body-calibrated retargeting, target-model forward kinematics and descriptor extraction to the same robot model. Windowed reference streaming then makes the resulting body-specific reference bank practical for joint policy training.

\section{Structured Embodiment-Conditioned Tracking}\label{sec:method}
We learn one feedback policy for heterogeneous humanoids. Its inputs distinguish what the robot should do, what it is currently doing and which body must execute the action. Motion references specify the target, observations describe the execution, and an embodiment descriptor specifies the executing body's structure, morphology and dynamics.

\subsection{A Shared Tracking Interface}
Each robot's native joints are bound to $J=38$ semantic slots. A presence mask $m_j$ identifies existing joints and an action mask $\mask_j$ identifies actuated channels. Missing slots are padded and excluded from masked operations; outputs for inactive channels are not applied to the robot. This common interface aligns observations, embodiment information and actions without replacing the native robot's dynamics.

Let $e$ denote an embodiment descriptor, $o_t$ the policy observations and $r_t^{e,\mathrm{win}}$ a temporal window of body-matched references. We learn
\begin{equation}
\begin{aligned}
&a_t\sim\pi_\theta(\cdot\mid o_t,r_t^{e,\mathrm{win}},\mask,h_g,\{h_j\}),\\
&(h_g,\{h_j\})=E_\psi(e).
\end{aligned}
\label{eq:policy}
\end{equation}
The encoder $E_\psi$ produces a global condition $h_g$ and joint-aligned conditions $h_j$. The action $a_t\in\mathbb R^J$ represents normalized joint-target offsets, converted to native position commands through joint binding, action scaling and PD control. The policy predicts executable control actions, not a new reference motion or the next simulator state.

Observations combine state and action histories as $o_t=[o_t^{\mathrm{state}};a_t^{\mathrm{hist}}]$, where $a_t^{\mathrm{hist}}=[a_{t-1};a_{t-2};a_{t-3}]$. State observations include root angular velocity, projected gravity, joint positions and joint velocities at offsets $\{0,1,2,3,4,8,16\}$ policy steps. The reference input contains root-position, root-orientation and joint-position targets at offsets $\{-8,-4,-2,0,1,2,3,4\}$. At 50~Hz, the state history spans 0.32~s and reference lookahead reaches 0.08~s. We use \emph{observations} as the collective term; history is an explicit part of this input.

\subsection{A Descriptor of the Executing Body}
We extract a structured embodiment descriptor $e=(G,k,d)$ from the robot model and its actuator configuration. The descriptor contains measured or specified model properties, not learned features or a robot identity label.

\paragraph{Structure}
The graph $G=(\mathcal V,\mathcal E,\ell)$ records present semantic joints, their parent--child topology and discrete attributes such as joint type, depth, side and semantic role. Presence and action masks distinguish missing joints from existing but inactive channels; the present-joint fraction summarizes joint presence at the whole-body level.

\paragraph{Morphology}
Morphology is $k=(k_g,\{k_j\})$, where whole-body geometry $k_g$ includes the root-to-support offset and root-frame collision bounds. Joint-level geometry is
\begin{equation}
k_j=[q_j^0,q_j^-,q_j^+,u_j,t_j,\operatorname{rot6D}(R_j)],
\label{eq:morphology}
\end{equation}
where $q_j^0$ is the nominal angle, $q_j^-,q_j^+$ are limits, $u_j$ is the joint axis, and $(t_j,R_j)$ is the parent-relative transform. Rotations use continuous six-dimensional coordinates \cite{zhou2019rotation}.

\paragraph{Dynamics}
Dynamics is $d=(d_g,\{d_j\})$. Whole-body fields $d_g$ describe total mass, center of mass and inertia about that center. Joint-level fields $d_j$ contain segment mass, center of mass and inertia in the joint frame, PD gains, effort and velocity limits, reflected armature and friction. These values describe the generated model actually used for reference construction and simulation, including its variant-specific scaling.

\subsection{From Descriptors to Body Conditions}
Group-specific projections form one global token and one token per semantic joint. The Graph encoder first exchanges information along parent--child connections and through the global token; subsequent blocks integrate all present tokens. Let $Z_0$ and $Z_j$ denote its contextualized global and joint tokens. Learned readouts produce
\begin{equation}
\begin{aligned}
&h_g=\rho_g\!\left([Z_0;\operatorname{mean}_{j:m_j=1}Z_j]\right),\\
&h_j=\rho_j(Z_j),\qquad j\text{ with }m_j=1.
\end{aligned}
\label{eq:readout}
\end{equation}
The global readout combines whole-body context with pooled joint information; the joint readout preserves action-channel correspondence, with absent slots zeroed. Absolute and normalized relative views provide complementary descriptions of body properties. The encoder tokens have width 128, while $h_g\in\mathbb R^{64}$ and $h_j\in\mathbb R^{16}$ are the conditions delivered to the policy.

We distinguish three levels of representation: the \emph{descriptor} is the structured model information, \emph{tokens} are intermediate learned features, and \emph{body conditions} are the readouts supplied to the policy. Flat-MLP replaces graph-based processing with a dense mapping of the same continuous fields, categorical metadata and masks. It retains the common output interface and Concat conditioning. Thus, Flat-MLP is an encoder ablation within \method{}, not a separate tracking framework or a reduced descriptor.

\begin{figure*}[!t]
\centering\includegraphics[width=\textwidth]{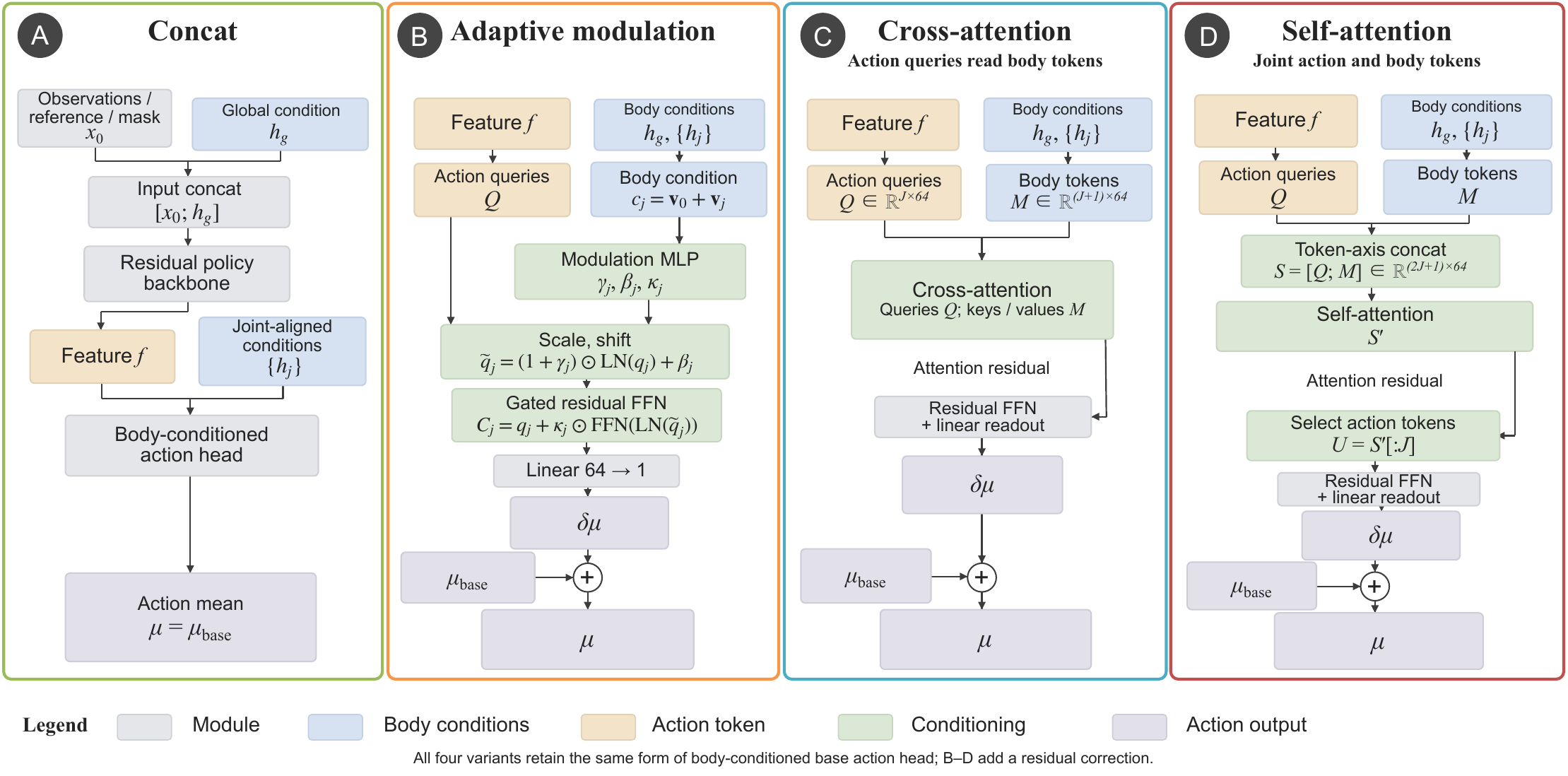}
\caption{\textbf{Four embodiment-conditioning mechanisms.} All variants use the same Graph encoder architecture and body-conditioned base action-head design. Concat supplies $h_g$ to the feedback backbone; the other three add residual action corrections. The interaction branches use $J=38$ action slots and width-64 tokens, distinct from the encoder's width-128 tokens.}
\label{fig:fusion}
\end{figure*}

\subsection{How Embodiment Information Enters Control}
The shared feedback backbone combines base and residual branches,
\begin{equation}
f=B(x)+\sigma(\alpha)R(x),
\end{equation}
where $B$ and $R$ are base and residual multilayer perceptrons, $\alpha$ is a learned gate and $\sigma$ is the sigmoid. With $x_0=[o_t;r_t^{e,\mathrm{win}};\mask]$, Concat uses $x=[x_0;h_g]$; the other mechanisms use $x=x_0$ and add an interaction branch. All four retain a joint-conditioned base action head,
\begin{equation}
\mu_j^{\mathrm{base}}=W_{o,j}f+b_j+
\eta_j\langle W_qf,h_j\rangle/\sqrt{D_h},
\label{eq:head}
\end{equation}
where $W_{o,j},b_j$ define the linear action readout, $W_q$ projects $f$ to dimension $D_h=16$ and $\eta_j$ is a learned gate initialized to zero. The final term links each output to the corresponding joint's properties.

Figure~\ref{fig:fusion} compares four embodiment-conditioning mechanisms. \emph{Concat} provides global conditions before feature extraction. \emph{Adaptive modulation} combines projected global and joint conditions as $c_j=v_0+v_j$ to generate action-query scale, shift and gate parameters. \emph{Cross-attention} lets action queries read body tokens. \emph{Self-attention} jointly processes action and body tokens \cite{perez2018film,peebles2023dit,vaswani2017attention}. The latter three use parameter-matched interaction modules and predict residual corrections,
\begin{equation}
\mu=\mu^{\mathrm{base}}+\Delta\mu,\qquad
\Delta\mu=F_{\mathrm{cond}}(f,h_g,\{h_j\},\mask).
\label{eq:conditioning-residual}
\end{equation}
Concat uses $\mu=\mu^{\mathrm{base}}$. Encoder capacity is examined separately using Graph and the two Flat-MLP variants.

\subsection{Joint Policy and Representation Learning}
PPO optimizes tracking return over sampled bodies and references \cite{schulman2017ppo}. We jointly train the encoder and the policy, with auxiliary objectives for descriptor reconstruction, paired-view agreement, corruption consistency and regularization of feature variance and covariance \cite{bardes2022vicreg}:
\begin{equation}
\begin{aligned}
\mathcal L_{\mathrm{aux}}={}&\lambda_{\mathrm{rec}}\mathcal L_{\mathrm{rec}}
+\lambda_{\mathrm{view}}\mathcal L_{\mathrm{view}}\\
&+\lambda_{\mathrm{corr}}\mathcal L_{\mathrm{corr}}
+\lambda_{\mathrm{vc}}(\mathcal L_{\mathrm{var}}+\mathcal L_{\mathrm{cov}}).
\end{aligned}
\label{eq:auxiliary}
\end{equation}
These objectives encourage the conditions to retain embodiment information while remaining useful for control. The asymmetric critic receives privileged observations and detached body conditions, so value-loss gradients do not update the encoder. Architecture, loss weights and optimization settings are detailed in Appendix~\ref{app:implementation}.

\section{Consistent Cross-Embodiment Learning}\label{sec:infrastructure}
We build a cross-embodiment learning infrastructure that keeps randomized robot models, embodiment descriptors and motion references consistent throughout training. An efficient heterogeneous runtime brings these components together to train one shared policy across thousands of body variants.

\subsection{Coupled Body Models and Motion References}
We generate body variants from 16 humanoid templates, retaining the correspondence between each model, its actuator settings, its descriptor and its motion references. Morphological and dynamical properties are coupled during randomization so that the simulated body and its descriptor remain consistent.

Body-calibrated retargeting establishes motion correspondence for the target robot, extending the human-humanoid-tools workflow and its inverse-kinematics foundation \cite{human_humanoid_tools2026,nvidia2026somaretargeter}. Target-model forward kinematics then produces reference body poses and velocities. This pipeline turns each source motion into body-specific references that reflect the target robot's kinematics.

The resulting bank contains joint states, body poses and velocities indexed by body and motion. For the main training population, 4,096 variants and 64 LAFAN1 takes yield approximately 262K body--motion trajectories. This construction keeps each motion reference consistent with the kinematics of its target robot.

\subsection{Heterogeneous Simulation with a Shared Policy}
Our heterogeneous runtime simulates robot families with different structures, morphologies and dynamics within the same training process. The semantic interface maps their native states and control channels into shared policy tensors, while each environment retains its own robot model and actuator response. By combining rollouts from these bodies in each policy update, we train a single controller jointly across heterogeneous humanoids.

\subsection{Reference Streaming and Encoding Reuse}
Storing all body-specific references on the GPU would make memory grow with the complete body--motion bank. We instead keep the packed bank in host storage and stream active temporal windows to the GPU. Window boundaries preserve the reference sequence and its required temporal offsets. GPU-resident reference storage consequently depends on the active windows rather than the full library.

Embodiment descriptors are static across steps and many rollout samples refer to the same body. During rollout, the runtime reuses cached body conditions while encoder parameters are fixed. During optimization, it gathers unique body descriptors, computes their conditions with gradients and expands them to the associated samples. The cache is invalidated when the encoder changes. This reduces repeated work without freezing the learned representation.

Together, reference streaming and body-wise encoding reuse reduce GPU memory demand and redundant computation. With these designs, we jointly train one policy across thousands of body variants.

\section{A Cross-Embodiment Tracking Benchmark}\label{sec:benchmark}
\subsection{Separating Motion and Body Generalization}
The benchmark contains 464 bodies from 16 seen and five unseen humanoid families. Five strata distinguish sampled training bodies, new bodies within or outside the training randomization support and new robot families (Table~\ref{tab:benchmark}). For an unseen family, within/outside support is defined relative to that family's generator, not membership in the training distribution. In particular, SO tests a shift within a known family, whereas UI/UO test unseen-family transfer.

\begin{table}[!t]
\caption{Benchmark bodies and evaluation settings. Each seen stratum contains eight bodies per family; each unseen stratum contains eight bodies per unseen family.}\label{tab:benchmark}
\centering\small
\begin{tabular}{@{}p{.12\columnwidth}p{.72\columnwidth}r@{}}
\toprule Stratum & Body status & $N$\\\midrule
SS & Seen family, sampled training body &128\\
SI & Seen family, new within-support body &128\\
SO & Seen family, outside-support body &128\\
UI & Unseen family, within-parent support &40\\
UO & Unseen family, outside-parent support &40\\\midrule
\multicolumn{3}{@{}l}{Physics / policy: 200 / 50 Hz}\\
\multicolumn{3}{@{}l}{Evaluation horizons: 10 / 20 / 30 s}\\
\multicolumn{3}{@{}l}{Body-local error limits: 0.8 m / 1.6 rad}\\
\multicolumn{3}{@{}l}{Persistence: five policy steps (0.1 s)}\\
\multicolumn{3}{@{}l}{Action delay: 10 ms; new-action weight: 0.9}\\
\bottomrule
\end{tabular}
\end{table}

The main policies are trained on 64 LAFAN1 sequences. We evaluate motion generalization on eleven held-out, non-obstacle LAFAN1 sequences and eight sequences from 100STYLE, which is not used for training \cite{harvey2020lafan,mason2022style}. All methods use the same starting points and nested 10/20/30-s evaluation windows. Seven of the eight 100STYLE sequences are long enough for the 30-s evaluation.

\subsection{Control and Scoring}
Evaluation uses policy-mean actions, training-source normalization, native action conventions, fixed simulation settings and no online randomization. Action delay is two physics steps, with smoothing weight 0.9 on the new command.

Motion-tracking success (SR) means completing the requested window without five consecutive policy steps at or above either body-local error limit: 0.8~m in position or 1.6~rad in orientation. The local frame removes horizontal root translation and heading while retaining height and tilt; each error is the maximum over the tracked bodies. We report world-space tracking accuracy separately. Effective tracking time is
\begin{equation}
T_{\mathrm{track}}=n_{\mathrm{valid}}/50,
\end{equation}
where $n_{\mathrm{valid}}$ counts steps before the first failure, capped at the evaluation horizon. G-MPJPE measures world-space body-position error; RPE measures root-position error. L-MPJPE subtracts each trajectory's root translation without rotational alignment and therefore differs from the termination frame. Errors exclude the terminal sample and use each policy's valid tracking prefix.

For cross-embodiment tables, metrics are averaged within each family and then across families in the stratum. Paired interventions report pooled successes over paired trials. No-fall survival is a separate physical outcome, defined by sustained unintended upper-body ground contact; intended contacts in the reference are exempted. Its precise definition is given in Appendix~\ref{app:scoring}.

\subsection{Controlled Comparisons}\label{sec:comparison-protocol}
The fixed-budget comparison uses eight GPUs, 8,192 environments per GPU and 10,500 PPO updates of 32 steps per environment. All ten configurations use the same body population and reference bank. Concat denotes Graph + Concat unless an alternative encoder is specified.

We report training cost in PRO 6000D-equivalent GPU-hours, normalizing GPU time across hardware.

G1 comparisons separate the nominal robot from 24 randomized G1 bodies, with eight bodies per SS/SI/SO stratum. The specialist is trained only on nominal G1; these stratum labels refer to the shared policy's population. We show both a 4,096-environment specialist and a specialist with the shared policy's total $8\times8192$ environment budget.

For condition interventions and nominal-G1 specialization, we use a fixed, post-hoc LAFAN1-8 subset excluding three ground and get-up sequences. The excluded sequences are listed in Appendix~\ref{app:scoring}. We report this focused analysis alongside the full benchmark, with checkpoints, starts and scoring unchanged. Continuation and specialization budgets are reported separately from the fixed-budget main comparison.

\begin{table*}[!t]
\centering\normalsize
\caption{\normalfont Thirty-second family-balanced motion-tracking success (SR, $\%\!\uparrow$) across 464 bodies; stratum populations follow Table~\ref{tab:benchmark}. All ten shared-policy configurations are \method{} variants. The two Flat-MLP encoders use the same descriptor fields and Concat conditioning: 2.638M parameters for widths 256/256 and 0.469M for 42/42, versus 0.466M for Graph. Cost: PRO 6000D-equivalent GPU-hours at 10,500 updates; Flat-MLP (42) uses a throughput-calibrated estimate. --- denotes unreported values. Bold marks the best SR in each stratum. ProtoMotions covers nominal G1 only.}\label{tab:main}
\setlength{\tabcolsep}{2.7pt}\renewcommand{\arraystretch}{1.10}\begin{tabular*}{\textwidth}{@{\extracolsep{\fill}}lrrrrrrrrrrr@{}}
\toprule
 & \multicolumn{5}{c}{LAFAN1-11: SR $\uparrow$} & \multicolumn{5}{c}{100STYLE-8: SR $\uparrow$} & Cost $\downarrow$\\
\cmidrule(lr){2-6}\cmidrule(lr){7-11}
Configuration & SS & SI & SO & UI & UO & SS & SI & SO & UI & UO & GPU-h\\\midrule
\multicolumn{12}{l}{\textit{\method{}: descriptor and encoder ablations}}\\
Mask-only & 64.8 & 55.3 & 55.0 & 0.7 & 4.8 & 93.8 & 90.0 & 86.6 & 13.2 & 10.4 & 82.2\\
Identity & 65.3 & 56.0 & 54.9 & \textbf{5.9} & \textbf{8.2} & 95.1 & 88.7 & 87.7 & \textbf{21.4} & \textbf{28.2} & 80.8\\
Dynamics & 67.5 & 54.6 & 55.5 & 4.1 & 5.5 & 95.9 & 91.0 & 89.0 & 6.8 & 11.4 & 109.7\\
Structure + morphology & 64.8 & 55.0 & 53.6 & 1.1 & 1.6 & 96.5 & 90.7 & 89.7 & 8.6 & 6.1 & 111.3\\
Flat-MLP (42) + Concat & 67.4 & 56.0 & \textbf{56.9} & 1.8 & 5.0 & 97.5 & 92.7 & 91.4 & 5.0 & 14.6 & 92.7\\
\highlightoursrow
Flat-MLP (256) + Concat & \textbf{67.6} & 56.2 & 56.2 & 3.6 & 6.4 & \textbf{98.1} & \textbf{94.2} & \textbf{92.3} & 12.9 & 23.2 & 96.9\\
\midrule
\multicolumn{12}{l}{\textit{\method{}: full descriptor, Graph encoder, four conditioning mechanisms}}\\
\highlighttablerow{concatrowbg}
Graph + Concat & 67.3 & 55.7 & 56.1 & 0.5 & 0.9 & 97.1 & 92.3 & 91.9 & 2.5 & 6.1 & 111.2\\
Adaptive modulation & 67.4 & \textbf{56.7} & 56.6 & 0.0 & 2.5 & 97.1 & 92.2 & 90.4 & 6.1 & 7.5 & 144.3\\
Cross-attention & 62.6 & 53.6 & 52.4 & 0.5 & 1.6 & 93.2 & 88.2 & 85.8 & 12.1 & 13.6 & 151.2\\
\highlighttablerow{attentionrowbg}
Self-attention & 61.4 & 51.6 & 52.2 & 0.9 & 1.8 & 94.6 & 90.0 & 87.6 & 11.4 & 5.0 & 201.3\\
\midrule
\multicolumn{12}{l}{\textit{External tracker: official G1 only (not stratum averages)}}\\
ProtoMotions (G1 only) & \multicolumn{5}{c}{36.4\% (4/11)} & \multicolumn{5}{c}{85.7\% (6/7)} & ---\\
\bottomrule\end{tabular*}
\end{table*}

\section{Experiments}
We organize the experiments around three questions: whether shared training improves robustness to body variation, which embodiment information and conditioning mechanisms help control and where transfer breaks down. All shared-policy configurations, including Flat-MLP, are \method{} variants. The main comparison comprises ten configurations covering descriptor content, encoder architecture and conditioning mechanisms. Specialist comparisons use the same MimicLite stack; interventions and representation analyses use Self-attention.

\subsection{RQ1: Shared Tracking and Body Robustness}\label{sec:rq1}

\paragraph{One policy tracks across body variations}
On 100STYLE, Graph + Concat reaches 97.1/92.3/91.9\% 30-s tracking success rate (SR) on SS/SI/SO, compared with 93.8/90.0/86.6\% for Mask-only (Table~\ref{tab:main}). The 5.3-point SO gain shows that explicit body information remains useful outside the training randomization support within seen families. Flat-MLP (256) + Concat reaches 98.1/94.2/92.3\%. LAFAN1 is harder for all methods: Graph + Concat reaches 67.3/55.7/56.1\%.

Continuous rollouts across all 16 families and synchronized body-variant examples are shown in Appendix~\ref{app:qualitative}, Figs.~\ref{fig:qual-atlas-a}--\ref{fig:qual-bodies}.

\paragraph{Shared training builds robustness beyond single-body specialization}
We compare shared Graph + Concat with a nominal-G1 specialist, both trained for 10,500 updates using $8\times8192$ environments. On 24 randomized G1 bodies, 100STYLE SR increases from 42.9\% (72/168) to 99.4\% (167/168), despite the specialist assigning its entire environment budget to G1. On LAFAN1-8, completion rises from 28.1\% (54/192) to 93.8\% (180/192) and effective tracking time increases from 14.74 to 29.03~s. The full eleven-motion evaluation has the same ordering; per-stratum results are reported in Appendix~\ref{app:g1-results}, Table~\ref{tab:g1-body-robustness}. Shared training thus delivers substantially greater robustness to body variation at the same total environment-transition budget.

This benefit differs from nominal-body performance. Specialists remain competitive on standard G1, where no body shift is present. Across randomized G1 strata, shared training also reduces local and world-space errors relative to the specialist trained with the same number of environment transitions; detailed errors and nominal-body results appear in Appendix~\ref{app:g1-results}.

\begin{figure*}[!t]
\centering\includegraphics[width=.94\textwidth]{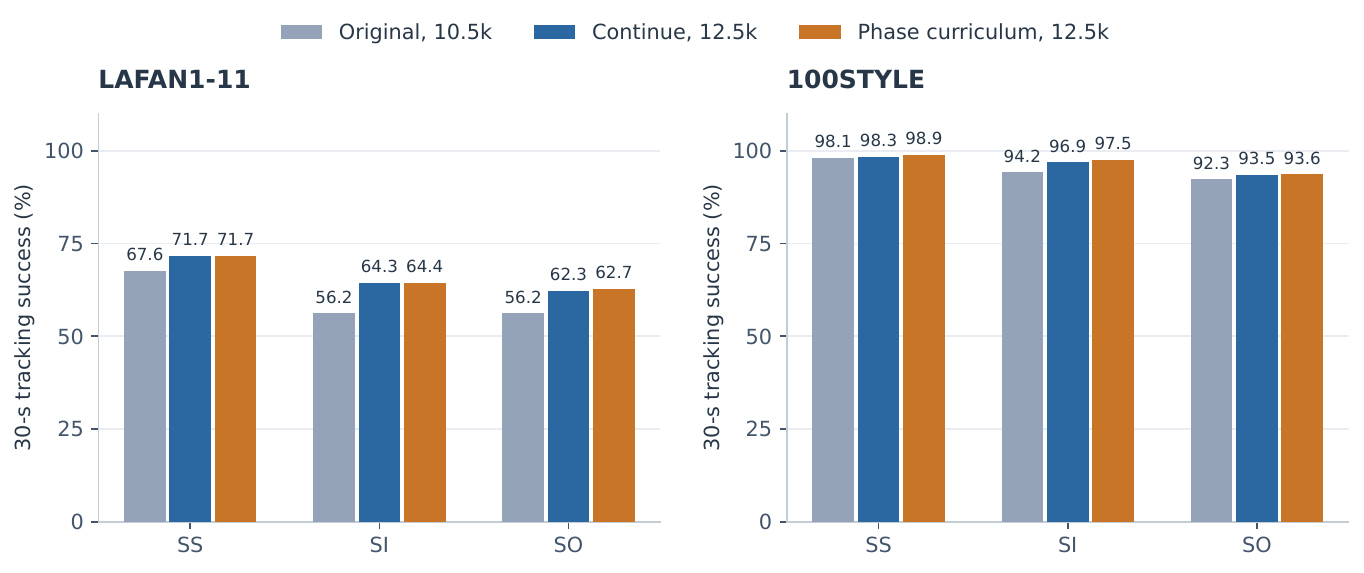}
\caption{\textbf{Continued training improves the shared tracker.} The same Flat-MLP + Concat checkpoint is continued for 2,000 updates on Train49 using ordinary or phase-based sampling. Bars show thirty-second family-balanced SR. Train49 excludes obstacle categories but retains ground and get-up motions.}
\label{fig:continuation}
\end{figure*}

\paragraph{Shared tracking improves with additional training}
We continue the Flat-MLP (256) + Concat checkpoint from 10,500 to 12,500 updates on Train49. This motion subset excludes obstacle categories but retains ground and get-up motions; the body population is unchanged. Ordinary continuation increases LAFAN1 SS/SI/SO SR from 67.6/56.2/56.2\% to 71.7/64.3/62.3\% (Fig.~\ref{fig:continuation}). The SI and SO gains are 8.1 and 6.1 percentage points. On 100STYLE, SR reaches 98.3/96.9/93.5\%.

Phase-based difficulty sampling yields 71.7/64.4/62.7\% on LAFAN1 and 98.9/97.5/93.6\% on 100STYLE. Both continuation variants improve on the original checkpoint across both suites, demonstrating that further training on Train49 effectively strengthens the shared tracker. Phase-based sampling provides an additional, smaller gain. Exact values and sampling scope are given in Appendix~\ref{app:continuation}.

\begin{table*}[!t]
\centering\normalsize
\caption{\normalfont Thirty-second tracking on SS/SI/SO (384 bodies, 16 families). Eff.: effective tracking time. G/L-MPJPE: global/root-relative position error; RPE: root-position error. Errors use valid tracking prefixes and are family-balanced. Bold marks the best value in each column.}\label{tab:quality}
\setlength{\tabcolsep}{2pt}\renewcommand{\arraystretch}{1.10}
\begin{tabular*}{\textwidth}{@{\extracolsep{\fill}}lrrrrrrrr@{}}
\toprule
 & \multicolumn{4}{c}{LAFAN1-11} & \multicolumn{4}{c}{100STYLE-8}\\
\cmidrule(lr){2-5}\cmidrule(lr){6-9}
Configuration & Eff. s $\uparrow$ & G-MPJPE & L-MPJPE & RPE & Eff. s $\uparrow$ & G-MPJPE & L-MPJPE & RPE\\
 & & \multicolumn{3}{c}{mm $\downarrow$} & & \multicolumn{3}{c}{mm $\downarrow$}\\\midrule
\multicolumn{9}{l}{\textit{\method{}: descriptor and encoder ablations}}\\
Mask-only & 21.5 & 958.2 & 71.4 & 955.9 & 27.7 & 1069.7 & 56.3 & 1064.5\\
Identity & 21.4 & 967.3 & 71.0 & 965.4 & 28.0 & 1103.2 & 56.5 & 1098.6\\
Dynamics & 21.6 & 815.9 & 69.5 & 812.8 & 28.3 & 767.4 & 53.1 & 761.6\\
Structure + morphology & 21.4 & 837.1 & 71.5 & 833.5 & 28.3 & 749.4 & 59.2 & 742.2\\
Flat-MLP (42) + Concat & 21.8 & 801.8 & 69.0 & 798.9 & 28.7 & 689.7 & \textbf{52.2} & 684.0\\
\highlightoursrow
Flat-MLP (256) + Concat & 21.8 & 793.5 & \textbf{68.6} & 790.4 & \textbf{28.8} & 734.8 & 52.8 & 729.5\\
\midrule
\multicolumn{9}{l}{\textit{\method{}: full descriptor, Graph encoder, four conditioning mechanisms}}\\
\highlighttablerow{concatrowbg}
Graph + Concat & 21.7 & 824.5 & 69.6 & 822.1 & 28.6 & 752.2 & 53.4 & 746.1\\
Adaptive modulation & \textbf{21.9} & 906.6 & 69.6 & 903.9 & 28.5 & 917.7 & 53.5 & 912.4\\
Cross-attention & 21.0 & 699.6 & 72.8 & 695.1 & 27.7 & 667.1 & 58.7 & 660.1\\
\highlighttablerow{attentionrowbg}
Self-attention & 20.7 & \textbf{660.7} & 74.2 & \textbf{656.0} & 27.9 & \textbf{643.7} & 59.4 & \textbf{636.5}\\
\bottomrule\end{tabular*}
\end{table*}

\subsection{RQ2: Embodiment Information and Its Use in Control}\label{sec:rq2}
\paragraph{Combining structure and morphology with dynamics improves tracking}
We group structure and morphology in one descriptor subset and dynamics in another and compare both with the full descriptor using Graph + Concat. The full descriptor reaches 91.9\% SR on 100STYLE SO, compared with 89.7\% for structure and morphology and 89.0\% for dynamics alone. The full descriptor's advantage over either subset is largest on SO among the seen-family strata. Combining these information groups therefore helps under larger within-family body shifts, with relative gains varying across suites and strata. Appendix~\ref{app:implementation} details the fields and structural encoding retained in each subset.

\paragraph{Strong completion does not require a large Flat-MLP encoder}
Flat-MLP (256) has 2,638,464 encoder parameters, compared with 466,048 for Graph. To control for encoder size, we evaluate Flat-MLP (42), with two hidden layers of width 42 and 469,360 encoder parameters, only 0.71\% more than Graph, while retaining the descriptor fields, output interface, Concat conditioning, data and update budget.

The smaller encoder reaches 67.4/56.0/56.9\% SR on LAFAN1 and 97.5/92.7/91.4\% on 100STYLE. Graph + Concat obtains 67.3/55.7/56.1\% and 97.1/92.3/91.9\%, respectively. At nearly the same encoder parameter budget, Flat-MLP and Graph achieve comparable completion rates. This result shows that a compact, simple encoder can make effective use of embodiment information for whole-body tracking. Parameter scope and the exact capacities are given in Appendix~\ref{app:capacity}.

\paragraph{Motion completion and global accuracy favor different choices}
We compare four conditioning mechanisms while keeping the descriptor and Graph encoder architecture fixed. Adaptive modulation has the highest LAFAN1 SI/SO SR among these four, whereas Concat leads on 100STYLE SI/SO. Self-attention has lower completion rates but the lowest measured G-MPJPE and RPE in both suites (Table~\ref{tab:quality}). Its G-MPJPE is 660.7/643.7~mm on LAFAN1/100STYLE, compared with 958.2/1069.7~mm for Mask-only. Flat-MLP variants instead have the lowest local errors. These prefix-based errors characterize tracking fidelity and complement SR and effective tracking time.

Reference overlays and root-position errors in Appendix~\ref{app:qual-drift} visualize the distinction between motion appearance and global tracking fidelity.

\begin{figure}[!t]
\centering\includegraphics[width=\columnwidth]{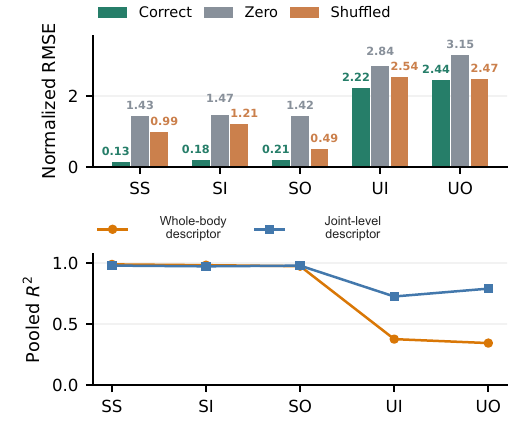}
\caption{\textbf{Retention of embodiment information.} Frozen encoder and reconstruction heads evaluated across five body strata. Whole-body descriptor reconstruction degrades more strongly on unseen families than joint-level descriptor reconstruction.}
\label{fig:reconstruction}
\end{figure}

\paragraph{Learned conditions retain embodiment information}
Frozen reconstruction heads recover descriptor fields from the learned conditions, achieving whole-body descriptor $R^2$ of 0.991/0.985/0.976 and joint-level descriptor $R^2$ of 0.981/0.975/0.981 on SS/SI/SO (Fig.~\ref{fig:reconstruction}). Correct conditions reduce reconstruction RMSE by 85.5--91.0\% there relative to Zero and outperform within-family shuffling. Reconstruction indicates what information is retained; the paired interventions separately test whether matching that information benefits control.

\begin{figure*}[!t]
\centering\includegraphics[width=\textwidth]{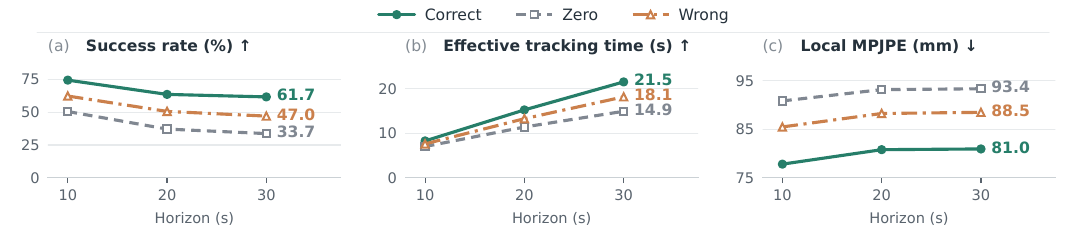}
\caption{\textbf{Body-matched conditions improve closed-loop tracking.} The same Self-attention checkpoint is evaluated on 464 bodies and eight LAFAN1 motions under Correct, Zero and Wrong conditions. Panels show trial-pooled SR, effective tracking time and root-relative L-MPJPE. Three repeats yield 11,136 trials per condition and horizon; exact success counts appear in Appendix~\ref{app:scoring}, Table~\ref{tab:conditions_lafan8}.}
\label{fig:intervention}
\end{figure*}

\paragraph{Matching conditions to the executing body matters}
We fix the checkpoint, body, reference and simulator, changing only the Self-attention policy's conditions. Correct retains body-matched conditions, Zero removes them, and Wrong substitutes another training parent's conditions selected by action-mask similarity. On the fixed LAFAN1-8 subset, thirty-second SR is \TokenSubsetCorrect{}\% for Correct, \TokenSubsetWrong{}\% for Wrong and \TokenSubsetZero{}\% for Zero (Fig.~\ref{fig:intervention}). The paired Correct--Wrong difference is \TokenSubsetWrongGain{} percentage points, with a body-cluster 95\% interval of [\TokenSubsetWrongLow{}, \TokenSubsetWrongHigh{}]. Effective tracking time and local position accuracy also improve. These controlled interventions demonstrate a direct closed-loop tracking benefit from matching the conditions to the executing body.

Paired Correct, Wrong and Zero rollouts in Appendix~\ref{app:qual-conditions} show how the condition interventions change actual execution under fixed body and motion inputs.

\begin{figure*}[!t]
\centering\includegraphics[width=.96\textwidth]{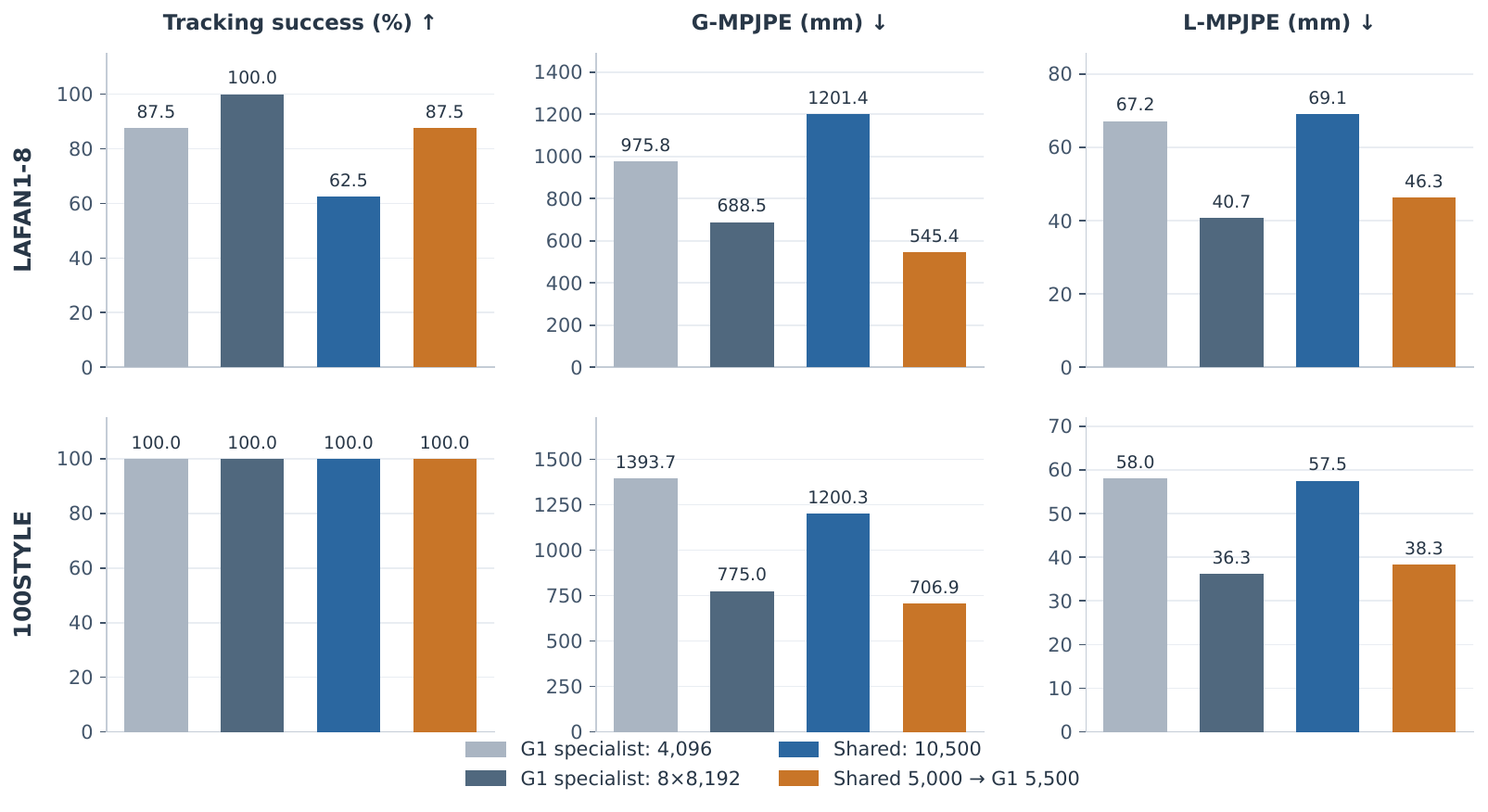}
\caption{\textbf{Shared pretraining followed by G1 specialization.} All policies use 10,500 updates; the two specialists differ in environment budget. Shared 5,000 $\rightarrow$ G1 5,500 retains Graph + Concat and improves nominal-G1 completion and world-space fidelity over shared-only training. LAFAN1-8 and 100STYLE have eight and seven eligible motions at 30~s. Full metrics appear in Appendix~\ref{app:g1-results}, Table~\ref{tab:g1-specialization}.}
\label{fig:g1-specialization}
\end{figure*}

\subsection{RQ3: Transfer Limits and Target Specialization}\label{sec:rq3}
\paragraph{Within-family robustness does not imply unseen-family transfer}
Across the ten configurations, LAFAN1 UI/UO SR remains at or below 8.2\%, whereas seen-family SR ranges from 51.6 to 67.6\%. Identity performs best on unseen families in the main SR table, so richer descriptors alone do not resolve this gap. Whole-body descriptor reconstruction $R^2$ drops to 0.377/0.344 on UI/UO, while joint-level descriptor reconstruction retains 0.727/0.792. This reconstruction gap suggests difficulty combining familiar joint properties into unfamiliar body structures, motivating broader family coverage alongside within-family randomization.

\paragraph{Specialization improves nominal-G1 tracking}
We compare four policies on standard G1: two single-body specialists, the shared Graph + Concat policy and a policy trained for 5,000 shared updates followed by 5,500 G1-only updates. The last two retain Graph + Concat throughout; all four have 10,500 total updates, with environment budgets reported explicitly.

On LAFAN1-8, specialization increases completion from 62.5\% (5/8) to 87.5\% (7/8), while G-MPJPE decreases from 1201.4 to 545.4~mm (Fig.~\ref{fig:g1-specialization}). On 100STYLE, both complete all seven eligible windows, but G-MPJPE falls from 1200.3 to 706.9~mm. The specialized policy has the lowest G-MPJPE and RPE among the four policies in both suites. The large single-body specialist still has higher LAFAN1 SR and lower L-MPJPE. Together, these results establish shared pretraining followed by target specialization as an effective route to improved nominal-G1 tracking, with particularly strong gains in global accuracy.

Because G1 is a seen training family, this experiment evaluates within-family target specialization. Appendix~\ref{app:g1-results} reports detailed nominal-G1 results and the external ProtoMotions comparison \cite{tessler2025protomotions}.
Synchronized crouched-walking examples in Appendix~\ref{app:qual-specialization} illustrate the corresponding changes in reference alignment and root-position error.

\section{Discussion and Conclusion}
We present \method{}, a scalable infrastructure and shared policy for embodiment-conditioned humanoid tracking. By connecting randomized robot models, descriptors, body-matched references and heterogeneous simulation, we enable joint training across thousands of body variants. Reference streaming and body-wise encoding reuse make this scale practical within a shared training runtime.

Our results demonstrate that explicit embodiment conditioning supports stable and robust whole-body tracking across diverse body variations within known humanoid families, even with a compact Flat-MLP encoder and simple concatenation. Self-attention provides the lowest measured world-space errors among the compared configurations, indicating that motion completion and global tracking accuracy favor different conditioning designs. Paired interventions establish the closed-loop benefit of body-matched conditions. Shared training substantially improves robustness over single-body specialization and target specialization further improves completion and tracking fidelity. Together, these results provide a practical foundation for scalable, embodiment-aware whole-body control.

The gap between within-family robustness and unseen-family transfer highlights a distinction between learning variations of familiar bodies and transferring control to new robot families. Broader structural coverage and contact-aware reference generation offer promising directions for extending the framework, together with obstacle-aware simulation and real-robot validation.

\FloatBarrier

\clearpage
\appendices
\section{Implementation and Training Details}\label{app:implementation}
\subsection{Inputs, Descriptor Fields and Network Interfaces}
The common interface has 38 semantic joint slots. Continuous descriptor fields comprise 18 whole-body values and 31 values per joint, totaling 1,196 before categorical fields and masks. The structure metadata records parent slot, depth, joint type, side and semantic role. Descriptor normalization uses training-source statistics and is not refitted on benchmark bodies. A record index selects the relevant descriptor; it is not passed to the encoder as a numerical body-identity feature.

The Graph encoder uses three attention blocks with width 128, four heads and feed-forward width 256. Its absolute and response readouts have dimensions 32 and 96. The policy adapter produces 64-dimensional global conditions and 16-dimensional joint conditions. The width-64 tokens in the conditioning modules are a separate projection, not the encoder's intermediate token width.

The Structure + morphology ablation retains geometric fields, structural metadata and parent--child attention connections, while removing dynamical fields. The Dynamics ablation retains dynamical fields but removes geometric fields and explicit structural embeddings; its first attention block attends to all present tokens rather than following parent--child connections. Both retain the semantic slots and presence and action masks required by the shared interface. The same field selection applies to relative descriptor views and auxiliary reconstruction targets.

Flat-MLP retains the continuous fields and uses fixed one-hot encodings for categorical metadata. Missing-joint blocks are zeroed, and masks remain explicit. It maps the flattened descriptor to the same global/joint representation interface, followed by the common pooling and readout components. Flat-MLP (256) and Flat-MLP (42) each have two dense hidden layers, with widths 256/256 and 42/42, respectively.

\subsection{Optimization}
Training trajectories and per-family learning progress are shown in Appendix~\ref{app:qual-training}.
Table~\ref{tab:optimization} records the fixed-budget training settings. The fixed-budget comparison uses one training run per configuration with seed 42. The feedback backbone has base widths 256/256/256 and residual widths 1024/1024/1024, with a sigmoid gate initialized from $\alpha=-4$. The critic has widths 1024/512/256 and receives detached body features. The optimizer is the MimicLite Muon/AdamW wrapper.

\begin{table}[h]
\centering\normalsize
\caption{Fixed-budget training configuration.}\label{tab:optimization}
\begin{tabular}{@{}lr@{}}
\toprule Setting & Value\\\midrule
Training environments & $8\times8192$\\
Policy updates / rollout steps & 10,500 / 32\\
PPO epochs / minibatches & 3 / 8\\
PPO clipping / gradient-norm cap & 0.2 / 1.0\\
Discount / GAE coefficient & 0.99 / 0.95\\
Initial actor learning rate & $3\times10^{-4}$\\
Desired KL, initial / late & 0.010 / 0.005\\
Entropy coefficient, initial / late & 0.012 / 0.001\\
Schedule transition & Updates 5,000--7,000\\
Training seed & 42\\
\bottomrule
\end{tabular}
\end{table}

For both Flat-MLP variants in the fixed-budget comparison, the encoder learning rate follows the actor's KL-adaptive rate rather than remaining fixed independently. Both use minibatch-mean KL control with a factor of 1.5, bounded actor rates of $10^{-5}$ to $10^{-2}$ and no epoch-level KL early stopping.

Representation losses use coefficients 0.05 for agreement between absolute and relative views, 0.02 for reconstruction, 0.02 for corruption consistency and 0.005 for variance and covariance regularization. Auxiliary sampling is capped at 2,048 unique body records. Corruption uses a drop probability of 0.05 and noise standard deviation 0.02. The critic's detached conditions prevent value gradients from changing the encoder.

\subsection{Rewards and Control}
The tracking objective combines root and body position, orientation, linear-velocity and angular-velocity terms with joint-position and joint-velocity tracking. The recorded weights are 0.5 for each root term, 1.0 for body position/orientation and 0.5 for body velocities and joint-state terms. Exponential-kernel scales are 0.3 for position, 0.4 for orientation, 1.0 for linear velocity, 2.5 for angular velocity, 0.25 for joint position and 2.5 for joint velocity. Regularization includes action rate, joint velocity, joint limits, self-collision, foot air time and survival. Reward and penalty signs are defined by their respective terms.

Training uses delayed and smoothed position commands, with delay sampled from one to three physics steps and new-action smoothing weight in $[0.8,1.0]$. Evaluation fixes these to two steps and 0.9. The environment advances at 200~Hz and the policy at 50~Hz. Individual robot models retain their own action scales, gains and actuator limits.

\section{Encoder Capacity Control}\label{app:capacity}
\begin{table}[h]
\centering\small
\caption{Encoder parameters, not total actor--critic parameters. Readout dimensions and Concat conditioning are retained.}\label{tab:encoder-capacity}
\begin{tabular}{@{}lrr@{}}
\toprule Encoder & Hidden widths & Parameters\\\midrule
Graph & Attention width 128 & 466,048\\
Flat-MLP (256) & 256/256 & 2,638,464\\
Flat-MLP (42) & 42/42 & 469,360\\
\bottomrule
\end{tabular}
\end{table}
Flat-MLP (42) has 0.71\% more encoder parameters than Graph. Dense flattening connects many input coordinates to all hidden units and reconstructs joint-aligned outputs through dense layers, whereas graph processing shares transformations across tokens. Parameter sharing allows Graph to use wider token features at a similar parameter count. This comparison examines the role of encoder capacity while retaining each architecture's computational structure. Flat-MLP (42) uses the same body population, reference bank, Train64 motion set, seed and 10,500-update budget as the other fixed-budget configurations. Its full 464-body benchmark is included in the main tables; Fig.~\ref{fig:qual-capacity-learning} compares the corresponding learning trajectories.

\section{Evaluation Definitions and Scope}\label{app:scoring}
Each body is tested on eleven LAFAN1 and eight 100STYLE clips. At 30~s, the eligible count is eleven and seven, respectively. Across 464 bodies and all three horizons this yields 25,984 evaluated windows per policy. The evaluation has no motion repetition or padding. Main-table averages give each family equal weight within a stratum; they are not a pooled average over the 16 seen and five unseen families.

The post-hoc LAFAN1-8 subset excludes fallAndGetUp3\_subject1, ground2\_subject2 and ground2\_subject3 from the eleven-motion set.

Let $\tau$ be the first declared tracking-failure step and $N_h$ the requested horizon in policy steps. The valid prefix contains the samples before $\min(\tau,N_h)$, with the terminal failure sample excluded. Effective time is the prefix length divided by 50. Global and local position errors are averaged over that prefix. L-MPJPE removes root translation but not heading, whereas the body-local termination test removes horizontal translation and heading while retaining height and tilt. These two notions of local coordinates should not be interchanged.

For a requested horizon of $h$ seconds, the completion fraction is
\begin{equation}
C_h=\min\!\left(\frac{T_{\mathrm{track}}}{h},1\right).
\label{eq:completion-fraction}
\end{equation}
It measures the fraction of the window tracked, rather than the binary success of the entire trial.

No-fall survival is measured over complete rollouts, including time after tracking failure, with no reset. A fall requires 0.2~s of unintended pelvis/spine/torso/head ground contact and clearance below 5~mm. A contact is treated as intended if the corresponding body in the reference motion is within 3~cm of the ground in a $\pm0.26$-s neighborhood. Recovery does not erase a fall. Invalid full-window observations are excluded from the survival denominator, not counted as surviving.

The no-fall survival rate is
\begin{equation}
S_{\mathrm{nofall}}=\frac{N_{\mathrm{nofall}}}{N_{\mathrm{eligible}}},
\label{eq:nofall-rate}
\end{equation}
where the numerator counts eligible full-window rollouts without a declared fall. This outcome is distinct from motion-tracking SR.

For paired condition interventions, all 464 bodies execute the same eight-motion subset under Correct, Zero and Wrong conditions, with three repeats and the same evaluation settings. Each condition/horizon has 11,136 trials. The Correct--Wrong interval resamples body clusters within families using 2,000 bootstrap draws; it quantifies variation over benchmark bodies, not variation over independently trained seeds. Reported differences are calculated before rounding: 61.7\% minus 47.0\% need not equal the rounded 14.6-point paired estimate.

\begin{table}[!t]
\centering\normalsize
\caption{Success rates and counts for the paired interventions in Fig.~\ref{fig:intervention}. All conditions use the same Self-attention checkpoint, 464 bodies, eight LAFAN1 motions and three repeats, yielding 11,136 trials per horizon.}
\label{tab:conditions_lafan8}
\setlength{\tabcolsep}{3pt}
\renewcommand{\arraystretch}{1.2}
\begin{tabular}{@{}lccc@{}}
\toprule
Horizon & Correct & Zero & Wrong \\
\midrule
10 s
& \shortstack{\textbf{74.4\%}\\8289/\TokenSubsetN{}}
& \shortstack{50.5\%\\5626/\TokenSubsetN{}}
& \shortstack{62.3\%\\6941/\TokenSubsetN{}} \\
20 s
& \shortstack{\textbf{63.6\%}\\7079/\TokenSubsetN{}}
& \shortstack{37.1\%\\4137/\TokenSubsetN{}}
& \shortstack{50.6\%\\5636/\TokenSubsetN{}} \\
30 s
& \shortstack{\textbf{\TokenSubsetCorrect{}\%}\\\TokenSubsetCorrectCount{}/\TokenSubsetN{}}
& \shortstack{\TokenSubsetZero{}\%\\\TokenSubsetZeroCount{}/\TokenSubsetN{}}
& \shortstack{\TokenSubsetWrong{}\%\\\TokenSubsetWrongCount{}/\TokenSubsetN{}} \\
\bottomrule
\end{tabular}
\end{table}

\begin{table}[!htb]
\centering\normalsize
\caption{\normalfont Official G1: (motion-tracking successes, no-fall survivors). Trials at 10/20/30~s are 11/11/11 for LAFAN1 and 8/8/7 for 100STYLE.}\label{tab:officialg1}
\setlength{\tabcolsep}{1pt}\renewcommand{\arraystretch}{1.10}
\begin{tabular}{@{}p{.34\columnwidth}*{6}{>{\centering\arraybackslash}p{\dimexpr.11\columnwidth-2\tabcolsep\relax}}@{}}
\toprule
 & \multicolumn{3}{c}{LAFAN1-11} & \multicolumn{3}{c}{100STYLE-8}\\
\cmidrule(lr){2-4}\cmidrule(lr){5-7}
Method & 10 s & 20 s & 30 s & 10 s & 20 s & 30 s\\\midrule
G1 (4096) & (9,9) & (7,8) & (7,8) & (8,8) & (8,8) & (7,7)\\
G1 ($8\times8192$) & (9,9) & (9,9) & (8,8) & (8,8) & (8,8) & (7,7)\\
ProtoMotions & (6,6) & (4,5) & (4,5) & (7,7) & (7,7) & (6,6)\\
\midrule
\multicolumn{7}{@{}p{\columnwidth}@{}}{\textit{Shared \method{} policy (ours)}}\\
\method{} (Concat) & (7,7) & (5,5) & (5,5) & (8,8) & (8,8) & (7,7)\\
\bottomrule\end{tabular}
\end{table}

\section{Continued Shared-Policy Training}\label{app:continuation}
\begin{table*}[!t]
\centering\small
\caption{Continued shared-policy training with Flat-MLP + Concat. All runs use the paper body and reference bank. Continuations start from the same 10,500-update checkpoint and add 2,000 updates on Train49, which excludes obstacle categories but retains ground and get-up motions. Entries are 30-s family-balanced SR (\%).}
\label{tab:continuation}
\setlength{\tabcolsep}{5pt}
\begin{tabular*}{\textwidth}{@{\extracolsep{\fill}}lrrrrrrr@{}}
\toprule & & \multicolumn{3}{c}{LAFAN1-11} & \multicolumn{3}{c}{100STYLE}\\
\cmidrule(lr){3-5}\cmidrule(lr){6-8}
Training & Total updates & SS & SI & SO & SS & SI & SO\\\midrule
Original policy & 10,500 & 67.6 & 56.2 & 56.2 & 98.1 & 94.2 & 92.3\\
Ordinary continuation & 12,500 & 71.7 & 64.3 & 62.3 & 98.3 & 96.9 & 93.5\\
Phase curriculum & 12,500 & 71.7 & 64.4 & 62.7 & 98.9 & 97.5 & 93.6\\
\bottomrule\end{tabular*}
\end{table*}

All continuation arms start from the same Flat-MLP (256) + Concat checkpoint at 10,500 updates and add 2,000 updates. Train49 removes obstacle categories from Train64 while retaining ground/get-up motions. The body models, descriptors and body-matched references remain unchanged.

The Phase variant emphasizes difficult temporal portions of the reference motions through its sampling curriculum, while retaining uniform family allocations. It provides a modest gain over ordinary continuation under the same family allocation. Both continuation settings combine further optimization with the Train49 motion subset.

\section{Nominal and Randomized G1 Results}\label{app:g1-results}
\begin{table*}[!t]
\centering\small
\caption{Thirty-second tracking on 24 randomized G1 bodies (eight per stratum).
Both policies use $8\times8192$ training environments and 10,500 updates.
LAFAN1-8 is the fixed subset defined in Sec.~\ref{sec:comparison-protocol}; LAFAN1-11 retains all motions.
Each stratum has 88, 64 and 56 trials for LAFAN1-11, LAFAN1-8 and 100STYLE, respectively.
Eff.: effective tracking time. Bold marks the better result within each motion set.}
\label{tab:g1-body-robustness}
\setlength{\tabcolsep}{3pt}\renewcommand{\arraystretch}{1.0}
\begin{tabular*}{\textwidth}{@{\extracolsep{\fill}}llrrrrrr@{}}
\toprule
 & & \multicolumn{2}{c}{SS} & \multicolumn{2}{c}{SI} & \multicolumn{2}{c}{SO}\\
\cmidrule(lr){3-4}\cmidrule(lr){5-6}\cmidrule(lr){7-8}
Motion set & Training policy & SR (\%) $\uparrow$ & Eff. s $\uparrow$ & SR (\%) $\uparrow$ & Eff. s $\uparrow$ & SR (\%) $\uparrow$ & Eff. s $\uparrow$\\
\midrule
LAFAN1-11 & Single-body G1 & 14.8 & 10.67 & 13.6 & 11.22 & 33.0 & 14.68\\
 & Shared 16-family (Concat) & \textbf{72.7} & \textbf{24.21} & \textbf{63.6} & \textbf{22.68} & \textbf{68.2} & \textbf{23.57}\\
\midrule
LAFAN1-8 & Single-body G1 & 20.3 & 12.68 & 18.8 & 13.60 & 45.3 & 17.94\\
 & Shared 16-family (Concat) & \textbf{100.0} & \textbf{30.00} & \textbf{87.5} & \textbf{28.08} & \textbf{93.8} & \textbf{29.00}\\
\midrule
100STYLE & Single-body G1 & 28.6 & 15.60 & 35.7 & 16.28 & 64.3 & 22.25\\
 & Shared 16-family (Concat) & \textbf{100.0} & \textbf{30.00} & \textbf{98.2} & \textbf{29.60} & \textbf{100.0} & \textbf{30.00}\\
\bottomrule\end{tabular*}\end{table*}

\begin{table*}[!t]
\centering\small
\caption{Thirty-second tracking on nominal G1. Shared pretraining and target specialization retain Graph + Concat throughout, with 5,000 shared updates followed by 5,500 G1 updates. All policies have 10,500 total updates; all except the 4,096-environment specialist use $8\times8192$ environments. LAFAN1-8 and 100STYLE have eight and seven eligible motions, respectively. Errors are in mm over valid tracking prefixes; best results within each suite are bold.}
\label{tab:g1-specialization}
\setlength{\tabcolsep}{4pt}
\begin{tabular*}{\textwidth}{@{\extracolsep{\fill}}llrrrrr@{}}
\toprule Motion set & Training & SR (\%) $\uparrow$ & Eff. s $\uparrow$ & G-MPJPE $\downarrow$ & L-MPJPE $\downarrow$ & RPE $\downarrow$\\\midrule
LAFAN1-8 & G1 specialist (4096) & 87.5 (7/8) & 28.43 & 975.8 & 67.2 & 975.2\\
 & G1 specialist ($8\times8192$) & \textbf{100.0} (8/8) & \textbf{30.00} & 688.5 & \textbf{40.7} & 685.3\\
 & Shared N16 Concat & 62.5 (5/8) & 23.00 & 1201.4 & 69.1 & 1198.5\\
 & Shared 5000 + G1 5500 & 87.5 (7/8) & 28.59 & \textbf{545.4} & 46.3 & \textbf{544.8}\\
\midrule
100STYLE & G1 specialist (4096) & \textbf{100.0} (7/7) & \textbf{30.00} & 1393.7 & 58.0 & 1392.7\\
 & G1 specialist ($8\times8192$) & \textbf{100.0} (7/7) & \textbf{30.00} & 775.0 & \textbf{36.3} & 769.9\\
 & Shared N16 Concat & \textbf{100.0} (7/7) & \textbf{30.00} & 1200.3 & 57.5 & 1192.7\\
 & Shared 5000 + G1 5500 & \textbf{100.0} (7/7) & \textbf{30.00} & \textbf{706.9} & 38.3 & \textbf{702.9}\\
\bottomrule\end{tabular*}
\end{table*}

\begin{table*}[!t]
\centering\small
\caption{Thirty-second 100STYLE tracking on nominal and randomized G1 bodies at 10,500 updates.
Seven eligible motions give seven nominal-body trials and 56 trials per randomized stratum.
The shared policy uses $8\times8192$ training environments; specialist budgets are shown explicitly.
G/L-MPJPE and RPE are in mm and use each trajectory's valid tracking prefix.
Bold marks the best result within each test-body group.}
\label{tab:g1-tracking-quality}
\setlength{\tabcolsep}{3pt}\renewcommand{\arraystretch}{1.0}
\begin{tabular*}{\textwidth}{@{\extracolsep{\fill}}llrrrrr@{}}
\toprule
Test bodies & Training policy & SR (\%) $\uparrow$ & Eff. s $\uparrow$ & G-MPJPE $\downarrow$ & L-MPJPE $\downarrow$ & RPE $\downarrow$\\
\midrule
Nominal G1 & Single-body G1 (4096) & \textbf{100.0} & \textbf{30.00} & 1393.7 & 58.0 & 1392.7\\
 & Single-body G1 ($8\times8192$) & \textbf{100.0} & \textbf{30.00} & \textbf{775.0} & \textbf{36.3} & \textbf{769.9}\\
 & Shared 16-family (Concat) & \textbf{100.0} & \textbf{30.00} & 1200.3 & 57.5 & 1192.7\\
\midrule
Randomized G1: SS & Single-body G1 ($8\times8192$) & 28.6 & 15.60 & 1060.9 & 72.6 & 1053.6\\
 & Shared 16-family (Concat) & \textbf{100.0} & \textbf{30.00} & \textbf{1043.7} & \textbf{47.1} & \textbf{1037.5}\\
\midrule
Randomized G1: SI & Single-body G1 ($8\times8192$) & 35.7 & 16.28 & 1460.2 & 78.2 & 1457.9\\
 & Shared 16-family (Concat) & \textbf{98.2} & \textbf{29.60} & \textbf{814.0} & \textbf{51.3} & \textbf{808.8}\\
\midrule
Randomized G1: SO & Single-body G1 ($8\times8192$) & 64.3 & 22.25 & 1782.8 & 66.6 & 1773.6\\
 & Shared 16-family (Concat) & \textbf{100.0} & \textbf{30.00} & \textbf{1621.2} & \textbf{49.5} & \textbf{1613.9}\\
\bottomrule\end{tabular*}\end{table*}

The 5,000+5,500 experiment keeps Graph + Concat throughout: the first stage trains across 16 families and the second uses nominal G1 only. Both stages use $8\times8192$ environments. This is distinct from continuing the Flat-MLP on Train49.

Tables~\ref{tab:g1-body-robustness}--\ref{tab:g1-tracking-quality} report per-stratum randomized-G1 robustness, nominal-G1 specialization and tracking precision for nominal and randomized bodies. Each table highlights the best values among its listed policies.

Table~\ref{tab:officialg1} compares nominal-G1 tracking and no-fall survival. ProtoMotions uses its original training data and compute budget, with native inputs and normalization under the benchmark physics and scoring.

\clearpage
\onecolumn
\section{Qualitative Results in Simulation}\label{app:qualitative}
\makeatletter
\newenvironment{qualplate}{\par\noindent\begin{minipage}{\textwidth}\def\@captype{figure}\centering}{\end{minipage}\par\vspace{2pt}}
\makeatother
\subsection{Shared Tracking Across Bodies and Motions}
The temporal composites below show recorded policy executions. Figure~\ref{fig:qual-spatial} retains world-space positions; the strips in Figs.~\ref{fig:qual-atlas-a} and~\ref{fig:qual-atlas-b} place successive poses side by side.
\begin{qualplate}
\includegraphics[width=\textwidth]{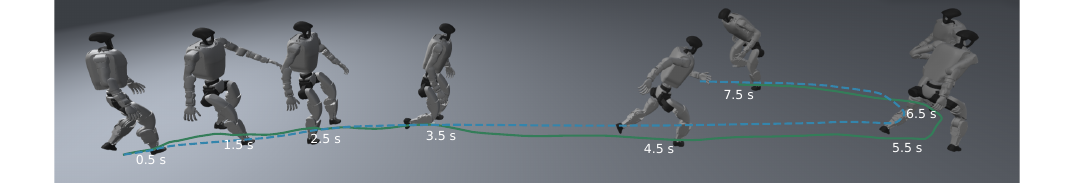}
\caption{\textbf{A trajectory expressed through successive body poses.} Eight states from one G1 running rollout, 0.5--7.5~s, share their original world frame. Solid green and dashed blue paths denote execution and reference.}
\label{fig:qual-spatial}
\end{qualplate}
\begin{qualplate}
\includegraphics[width=\textwidth]{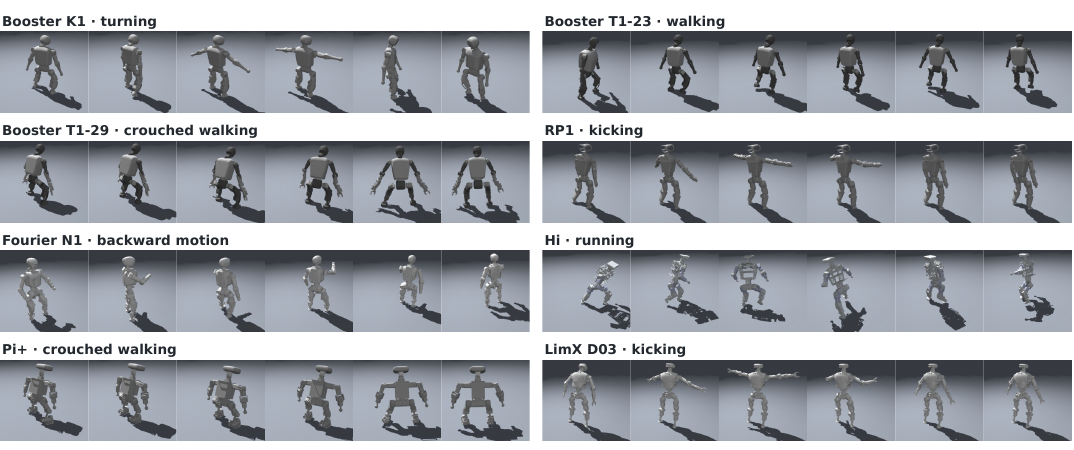}
\caption{\textbf{One shared policy across heterogeneous humanoids, part I.} Each row shows six states at 0.5, 1.2, 1.9, 2.6, 3.3 and 4.0~s. All families use the same Flat-MLP (256) + Concat checkpoint at 10,500 updates and held-out 100STYLE motions.}
\label{fig:qual-atlas-a}
\end{qualplate}
\begin{qualplate}
\includegraphics[width=\textwidth]{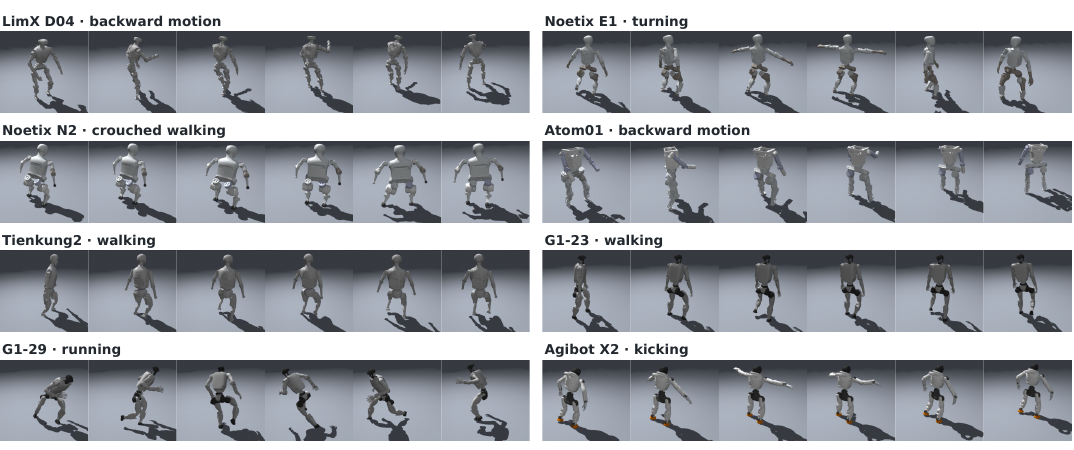}
\caption{\textbf{One shared policy across heterogeneous humanoids, part II.} The remaining eight families use the same checkpoint and times as Fig.~\ref{fig:qual-atlas-a}. Cameras are fixed within each strip and framed individually across families; native geometry and joint poses are retained.}
\label{fig:qual-atlas-b}
\end{qualplate}

\clearpage
\subsection{Body Variation and Reference Alignment}\label{app:qual-bodies}\label{app:qual-drift}
Opaque robots show execution and translucent cyan robots show body-matched references. Paired comparisons use identical camera settings and preserve the execution--reference offset at every displayed time.
\begin{qualplate}
\includegraphics[width=\textwidth]{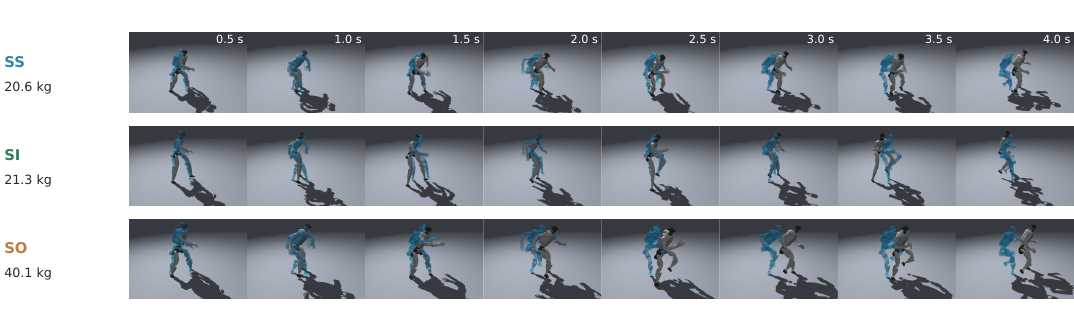}
\caption{\textbf{One motion across body variants.} Eight synchronized states per SS, SI and SO G1 body, all controlled by the same Self-attention policy. Model masses are 20.6, 21.3 and 40.1~kg; all three bodies retain 29 active joints.}
\label{fig:qual-bodies}
\end{qualplate}
\begin{qualplate}
\includegraphics[width=\textwidth]{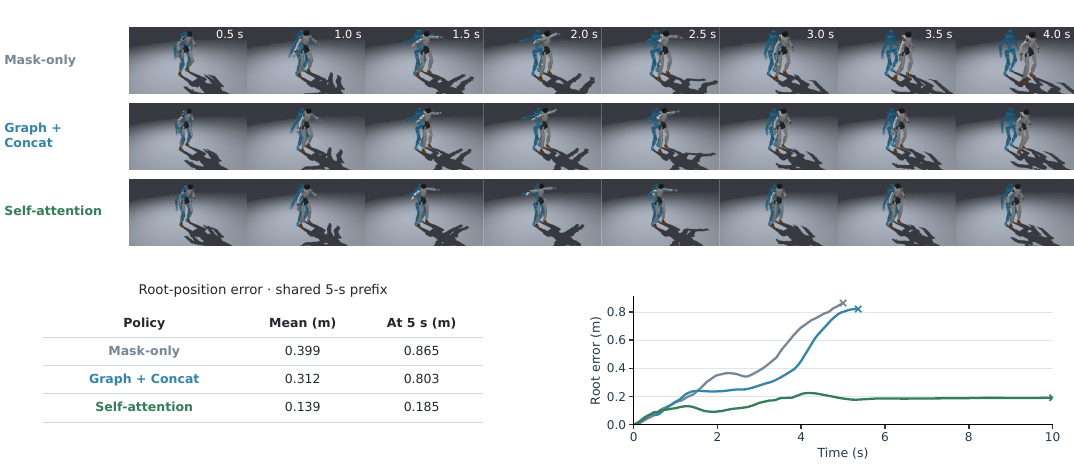}
\caption{\textbf{Pose fidelity and global displacement are complementary.} Eight synchronized states from an identical X2 kicking trial compare Mask-only, Graph + Concat and Self-attention. The table compares three-dimensional root error over the common five-second tracking prefix; curves retain each policy's complete valid prefix under the global-and-local criterion.}
\label{fig:qual-drift-x2}
\end{qualplate}
\begin{qualplate}
\includegraphics[width=\textwidth]{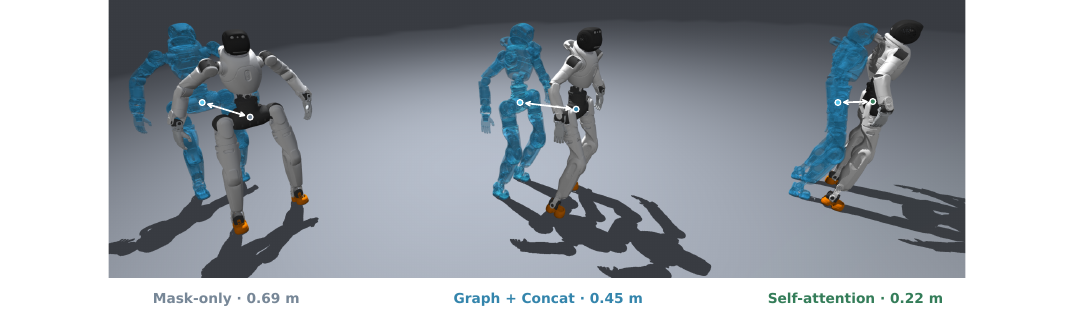}
\caption{\textbf{Reference alignment at 4.0~s.} Enlarged poses from Fig.~\ref{fig:qual-drift-x2} in a common presentation scene. Arrows connect executed and reference roots; labels report three-dimensional errors. Translating each pair for display preserves its relative offset.}
\label{fig:qual-detail}
\end{qualplate}

\clearpage
\subsection{Condition Interventions and Target Specialization}\label{app:qual-conditions}\label{app:qual-specialization}
Paired interventions hold the policy, body, reference and initialization fixed. Target specialization instead compares shared-only learning with a second training stage on nominal G1.
\begin{qualplate}
\includegraphics[width=\textwidth]{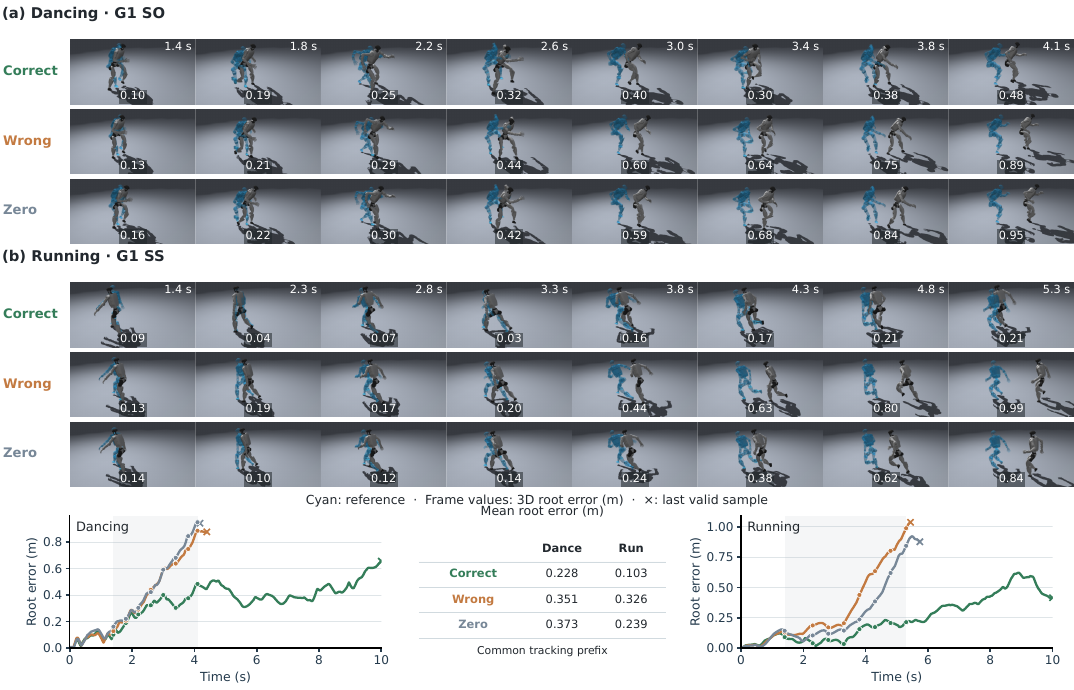}
\caption{\textbf{Correct, Wrong and Zero conditions across two motions.} Eight synchronized times per motion illustrate lower root error with Correct than with either alternative. These selected frames complement the aggregate interventions in Sec.~\ref{sec:rq2}. Curves retain the complete valid tracking prefixes; table means use common 4.18-s (dancing) and 5.44-s (running) prefixes. Cyan denotes reference poses, dots mark the displayed times and crosses mark the last valid samples.}
\label{fig:qual-condition-dance}
\end{qualplate}
\begin{qualplate}
\includegraphics[width=\textwidth]{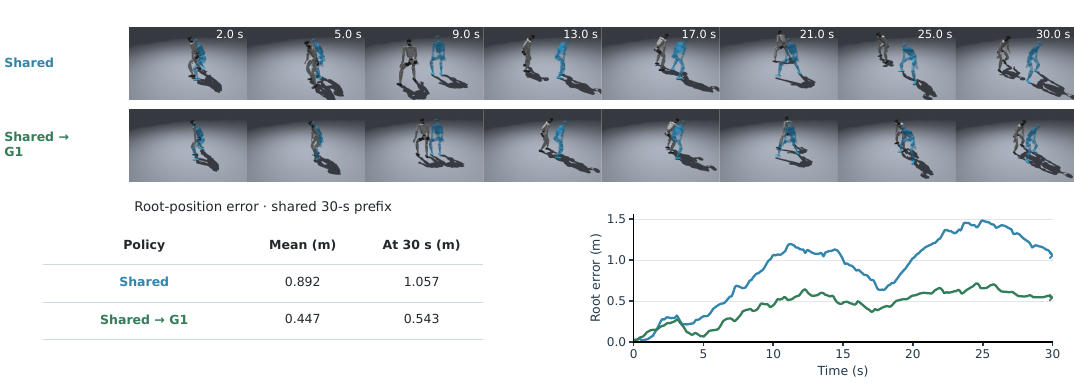}
\caption{\textbf{Specialization improves reference alignment during crouched walking.} Shared Graph + Concat training for 10,500 updates is compared with 5,000 shared plus 5,500 G1-only updates on the same thirty-second 100STYLE window. Eight synchronized times, root-error statistics and full curves complement the body-local evaluation in Table~\ref{tab:g1-specialization}.}
\label{fig:qual-specialization-crouch}
\end{qualplate}

\clearpage
\subsection{Learning Across Methods and Families}\label{app:qual-training}
\begin{figure}[!ht]
\centering\includegraphics[width=\textwidth]{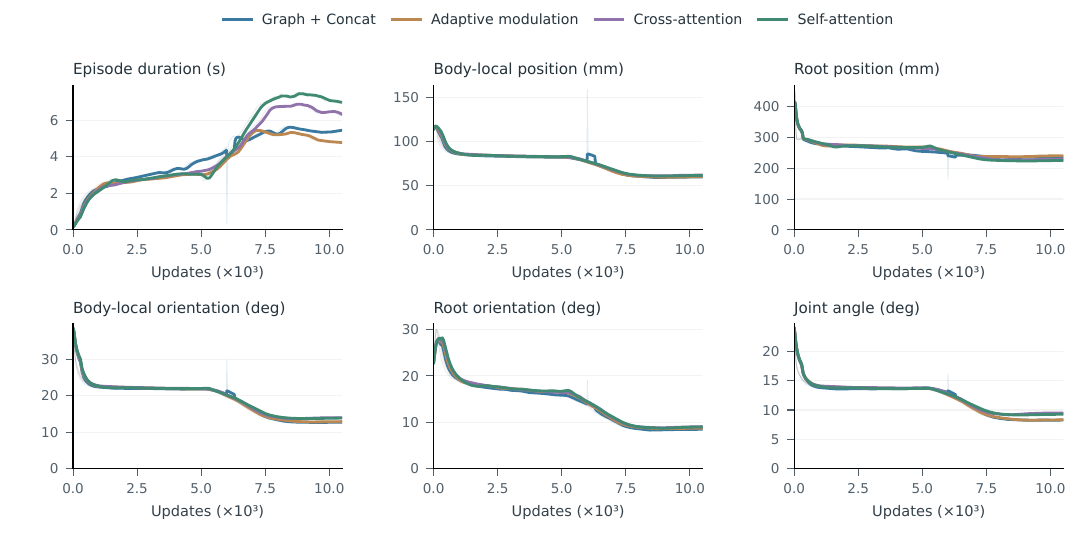}
\caption{\textbf{Tracking components during shared-policy learning.} Four conditioning mechanisms are compared through episode duration, body-local and root position, body-local and root orientation and joint-angle error. Each panel reports a family-balanced training statistic.}
\label{fig:qual-learning}
\end{figure}

\begin{figure}[!ht]
\centering\includegraphics[width=\textwidth]{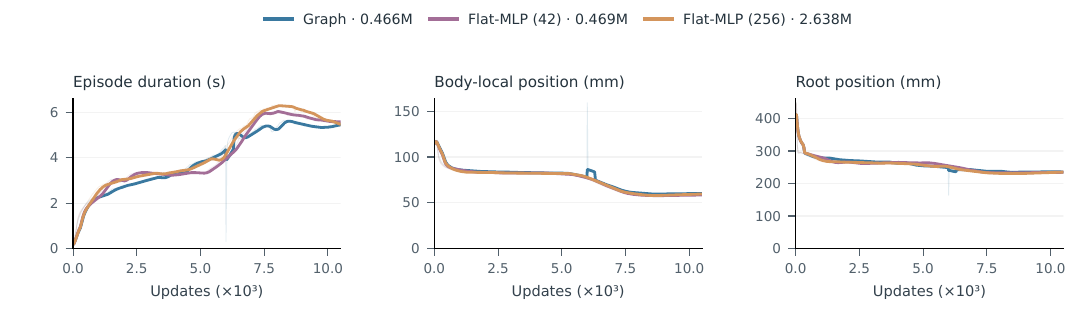}
\caption{\textbf{Encoder capacity and learning behavior.} Graph, Flat-MLP (42) and Flat-MLP (256) use the same descriptor fields and Concat conditioning. Legend values count encoder parameters; environment count, training seed and update budget are shared.}
\label{fig:qual-capacity-learning}
\end{figure}

\begin{figure}[!ht]
\centering\includegraphics[width=\textwidth]{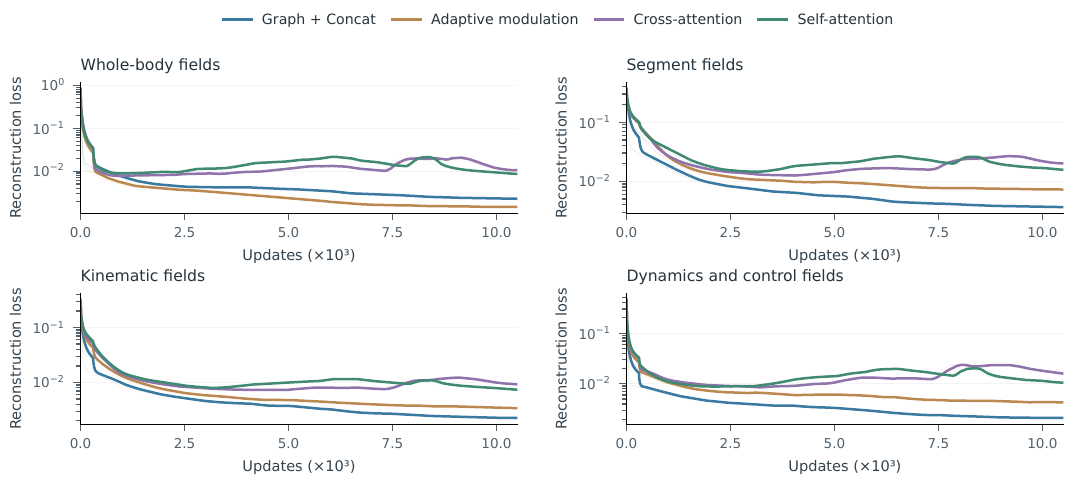}
\caption{\textbf{Learning to retain different descriptor fields.} Reconstruction objectives are shown separately for whole-body, segment, kinematic and dynamics and control fields. Values are the logged component losses on logarithmic axes, without rescaling across field groups.}
\label{fig:qual-descriptor-learning}
\end{figure}

\begin{figure}[!ht]
\centering\includegraphics[width=\textwidth]{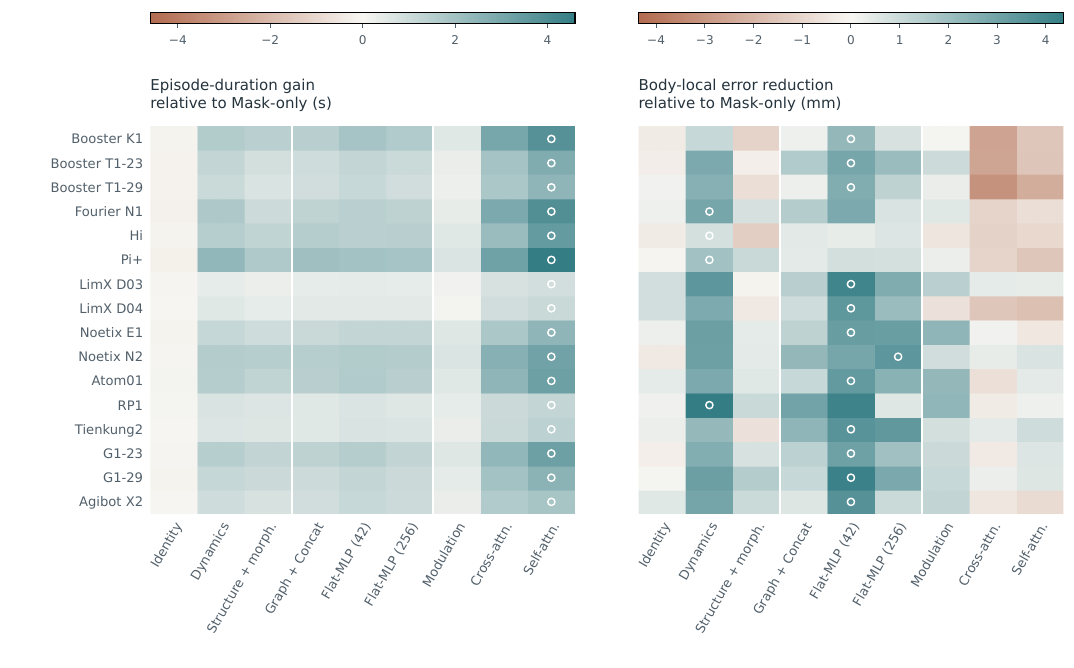}
\caption{\textbf{Training gains resolved by robot family.} Colors show final-300-update differences from Mask-only. Teal denotes longer episodes or lower body-local position error; orange denotes the reverse. Open circles identify the largest numerical improvement within each family.}
\label{fig:qual-family-learning}
\end{figure}

Figures~\ref{fig:qual-learning}--\ref{fig:qual-family-learning} examine complementary aspects of shared-policy learning: tracking components, encoder capacity, descriptor reconstruction and differences across families. All fixed-budget runs use seed 42 and $8\times8192$ environments. Tracking curves weight the sixteen families equally. Episode-summed position and angular errors are divided by episode length within each family before averaging; body-local errors use the training reference frame. Faint traces show logged samples and solid curves use trailing 300-update means.

Self-attention sustains longer training episodes, while local and global errors exhibit different orderings among conditioning mechanisms (Fig.~\ref{fig:qual-learning}). Descriptor-reconstruction curves further distinguish recovery of whole-body, segment, kinematic and dynamical fields (Fig.~\ref{fig:qual-descriptor-learning}). These views complement the held-out tracking and reconstruction evaluations.

The family-resolved comparison in Fig.~\ref{fig:qual-family-learning} shows that gains depend on both the executing body and the measured objective. Under Concat conditioning, the compact Flat-MLP follows learning trajectories close to those of the larger Flat-MLP and Graph encoders (Fig.~\ref{fig:qual-capacity-learning}), complementing the capacity control in Appendix~\ref{app:capacity}.

The radar in Fig.~\ref{fig:overview} compares Flat-MLP (256) + Concat with Mask-only on 100STYLE (30~s; 384 SS/SI/SO bodies). Scores are $\mathrm{SR}/\max(\mathrm{SR})$ or $\min(\mathrm{error})/\mathrm{error}$ across the two policies, on a shared 60--100\% scale; labels give raw values. Training-curve endpoints average the final 300 iterations.

\clearpage

\end{document}